\documentclass[final,3p,times,authoryear]{elsarticle} 

\usepackage{amssymb}
\usepackage{amsmath}
\usepackage{booktabs}
\usepackage{tikz}
\usetikzlibrary{calc}
\usetikzlibrary{math}

\usepackage{verbatim}

\journal{Information Fusion}

\journal{Information Fusion}

\begin{document}

\begin{frontmatter}



\title{Learning Image-based Tree Crown Segmentation from Enhanced Lidar-based Pseudo-labels}


\author[fgi,aalto,onera]{Julius Pesonen\corref{cor1}} 
\ead{julius.pesonen@nls.fi}
\cortext[cor1]{Corresponding author}
\author[fgi]{Stefan Rua}
\author[fgi]{Josef Taher}
\author[fgi]{Niko Koivumäki}
\author[fgi]{Xiaowei Yu}
\author[fgi]{Teemu Hakala}
\author[fgi]{Eija Honkavaara}

\affiliation[fgi]{organization={Department of Remote Sensing and Photogrammetry, Finnish Geospatial Research Institute},
            addressline={Vuorimiehentie 5}, 
            city={Espoo},
            postcode={02150}, 
            country={Finland}
}
\affiliation[aalto]{organization={Department of Computer Science, Aalto University},
            addressline={Otakaari 1B}, 
            city={Espoo},
            postcode={02150}, 
            country={Finland}
}
\affiliation[onera]{organization={DOTA, ONERA, Université de Toulouse},
city={Toulouse},
postcode={31000},
country={France}}

\begin{abstract}
Mapping individual tree crowns is essential for understanding and caring for our environment by improving the accuracy of urban tree inventories and precise monitoring of forest health. However, automatically separating tree crowns in aerial imagery is challenging because of factors like complex crown textures and frequent partial overlap between neighbouring trees. We present a method to train deep learning models to segment individual tree crowns from RGB and multispectral imagery, using pseudo-labels derived from aerial laser scanning (ALS) data. We show that incorporating information from a zero-shot instance segmentation model, Segment Anything Model 2 (SAM 2),  improves model training. Our method enables training domain-specific optical image-based models without requiring manual training annotation. We trained and evaluated the models with data from an urban area in Southern Finland and compared their performance with several baselines. Our best model achieved a mean intersection over union (mIoU) of 0.621 on manually delineated tree crowns in a geographically separate test area. This represents a 21\% increase in mIoU over the directly evaluated SAM 2-enhanced pseudo-labels (0.513) and a 19\% increase over the best image-based baseline (0.524). Incorporating SAM 2 into the pseudo-label generation process improved the best model mIoU by 9.9\%, from 0.565 to 0.621, while the average mIoU across the tested models increased by 8.4\%, from 0.557 to 0.604. These results highlight that automatically generated and foundation-model-enhanced pseudo-labels can provide effective domain-specific supervision for individual tree crown segmentation without manual training annotation, despite the noise in the initial ALS-based segments.
\end{abstract}


\end{frontmatter}

\color{black}

\section{Introduction}
Mapping individual tree crowns supports a wide range of environmental applications, including urban tree inventorying and forest health monitoring. High-resolution maps of urban trees can, for example, help assess their effects on the health and well-being of urban residents ~\citep{lai2019impact} and quantify their carbon storage~\citep{zhao2015quantifying,yang2025mapping}. In forest environments, individual-tree mapping is important not only for forest management~\citep{10.5849/forsci.12-134}, but also serves other important purposes, such as supporting biodiversity monitoring~\cite{ZHENG2022103074}, observing and mitigating increasing pest-related threats~\citep{hlasny2021bark,patacca2023significant}, and assessing forest fire risk~\citep{you2017geographical}.

Tree inventories can be conducted and maintained using various data sources, ranging from ground-based expert measurements~\citep{usda_urban_inv} to automated satellite image analysis~\citep{zhou2013mapping,zhang2023mapping}. Aerial data collection lies between these methods, offering quicker mapping of larger areas than field measurements and higher spatial resolution than satellite imaging. 
However, automatically separating individual trees from one another and from the background in aerial imagery data remains challenging, limiting the generation of high-resolution tree maps. In recent years, the best results have mostly been obtained using machine learning-based methods~\citep{zheng2024review}, but they typically require large amounts of human-annotated training data. The imagery used for tree detection can vary substantially across geographic locations, sensors, weather conditions, seasons, and even times of the day. Developing methods that perform reliably across such diverse conditions requires efficient methods for collecting representative training data.

The lack of large-scale datasets from varied geographic locations, sensors, and lighting scenarios has led to the development of mostly location-dependent models. 
These models are not robust enough to create generally reliable maps from a variety of locations, as our experiments highlight. 
Efficient training data generation methods offer a solution both to producing larger-scale datasets supporting the development of more robust models and to the efficient adaptation of models into new domains. 
One avenue to produce more training annotations for image-based models is through lidar-based data. 
Studies on the intersection of camera- and lidar-based systems on the task have, however, been limited. In some works, the fused sensor data were used directly to generate the tree shape estimates~\citep{favorskaya2015fusion,aubry2021multisensor}. 
In addition, using lidar-derived tree segments to train image-based models has been studied on data from the taiga-tundra ecotone with low tree density~\citep{lin2024individual} and in Germany with high-density lidar data~\citep{dersch2024semi}. 
However, neither of the previous studies provided comparisons between the proposed methods and other available learning-based methods that use optical image inputs.
A similar approach has also been used to train canopy height prediction models from satellite imagery~\citep{tolan2024very}, with a subsequent study further improving the spatial resolution of the predictions~\citep{brandt2026chmv2}.
Although these models provide high spatial resolution predictions, for satellite-based sensing, of up to 10~m in the earlier version and 1~m in the later version, the resolution remains far from the centimetre- to decimetre-scale achievable from both crewed and uncrewed aircraft.

Lidar data comes with the challenge of being less readily available than that of optical cameras due to the relative cost and rarity of the sensors. 
On the other hand, the best-performing methods using optical sensor data are typically based on supervised learning, meaning that they require, usually hand-crafted, labels to train. 
Various supervised models have been trained for the task on different datasets. Models which were publicly available at the time of this study, and thus available for testing on our data, have been described in Section~\ref{baselines}. 
The metric results obtained using such models vary heavily based on the data. The variation is caused by many factors, such as the resolution of the data, the similarity between test and training data, and the number of training samples.
It remains a major challenge to obtain sufficient annotated data to even evaluate the most generalising learning-based methods, as it is unfeasible to create hand-annotated datasets for all possible types of deployment scenarios.

As our method's performance approaches that of fully supervised models, we draw comparisons to the metrics reported in such studies.
A recent survey reports F1 scores ranging from 0.82 to 0.98, precision from 0.85 to 0.91, recall from 0.88 to 0.91, and mIoU from 0.49 to 0.94 across studies including LiDAR-based and semantic segmentation approaches ~\citep{wolk2024review}. For Detectree2, one of the baseline methods used in this study, reported test performance varies by area, with precision ranging from 0.60 to 0.71, recall from 0.54 to 0.66, and F1 scores from 0.57 to 0.69 ~\citep{ball2023accurate}. Similarly, a supervised U-Net, also used as a baseline in this study, achieved a precision of 0.71 and recall of 0.66 on aerial imagery, while performance on satellite imagery was slightly lower, with a precision of 0.63 and a recall of 0.64~\citep{freudenberg2022individual}.

Besides lidar data, leveraging information stored in pretrained image-based models can enable more robust learning, as has been proven by the consistent success of transfer learning across practically all of computer vision.
Specifically, the recent development of promptable image segmentation foundation models, such as the Segment Anything model (SAM)~\citep{kirillov2023segany}, or its more recent variants SAM 2 and 3~\citep{ravi2024sam2segmentimages,carion2025sam3segmentconcepts}, has led to new avenues to improve performance in various specific segmentation tasks by enabling new ways of extracting information without fine-tuning the models themselves.
The use of SAM for tree segmentation has been studied using learning-based prompts~\citep{chen2024rsprompter} or digital surface model-derived prompts~\citep{teng2025assessing}. SAM 2 has also been deployed on the task by combining it with the general tree detection model DeepForest~\citep{weinstein2020deepforest,chen2025zero}. Finally, Grounded SAM~\citep{ren2024grounded}, a combination of Grounding DINO~\citep{liu2023grounding} and Segment Anything Model, has been used for tree mapping in the context of disaster management~\citep{arnaudo2024fmars}. Different ways of prompting SAM 2, using either lidar, manual, or model-based prompts, have also been evaluated on their direct individual tree crown delineation performance~\citep{zhu2025leveraging}. 
Prior works have not suggested using automated tree crown segmentation data for targeting the domain transfer gap between different tree crown datasets, nor tested the methods against comparable image-based baselines. 

Our work analyses how local, automatically generated segmentation data can help us efficiently obtain location-specific models for individual tree crown segmentation.
In addition, we have evaluated how incorporating information from a more general segmentation foundation model can improve the performance of the task-specific downstream models. 
The methods were tested on datasets with a combination of high tree density and high spatial resolution, offering new, valuable insight into the effectiveness of the approach.
Most importantly, this work shows that tree crown segmentation models can be trained using RGB and multispectral imagery and canopy height model (CHM)-based segments derived from aerial laser scanning (ALS) surveys. 
We provide a comparison to alternative image-based methods, showing that training the image-based models, even with coarse or noisy labels, provides more robust results than the current zero-shot alternatives.
This presents a new solution for obtaining tree crown segments in areas where ALS surveys and aerial image collection have been conducted, as well as a new way to obtain training data for tree crown instance segmentation models, which can subsequently be deployed in areas where only image data is available. 
In addition, we highlight that incorporating information from an image-based foundation model enhances the predictions, an approach first introduced for this task in the master's thesis by~\citet{rua2025sam2}.
Our study extends the thesis with more detailed evaluation, analysis, and ablation.

\section{Materials and methods}

To study the possibility of using lidar-based pseudo-labels as automatically generated training references for a tree crown segmentation model, we first leveraged a dataset with the relevant requirements for training an image-based pseudo-supervised model. 
The trained model uses common hyperparameters besides the training data to ensure comparability with the relevant image-based baseline methods. 
We evaluated the proposed method using multiple datasets and compared the metrics to those available in the literature and the metrics the baseline models obtain in our local data.

\subsection{Study areas}
\label{sec:areas}

We studied tree crown segmentation using five datasets from four geographically distinct locations.
First off, most of the experiments were done using data from an urban area in the Espoonlahti district of Espoo, Finland (centred at approximately $60.1462^\circ\text{N}$ and $24.6587^\circ\text{E}$).
The location is characterised by its diversity of forest types, ranging from typical Fennoscandian coastal landscape to managed Finnish boreal forests. 
Based on the statistics from the study by~\citet{taher2026multispectral}, in the region, the tree species distribution consisted of approximately 37\% pine (Pinus sylvestris), 25\% birch (Betula sp.), 11\% maple (Acer platanoides), 11\% spruce (Picea sp.), 8\% aspen (Populus tremula), 3\% rowan (Sorbus sp.), 3\% linden (Tilia sp.), 2\% alder (Alnus sp.), and 1\% Oak (Quercus robur).

We used a separate test set captured from Sundsvall, Sweden (centred at approximately 63.0842$^\circ$N and 16.9217$^\circ$E), for testing how well the proposed models could perform in a scenario with a slight distribution shift. 
The area consisted of managed forest with a total of 3~562 referenced trees. 1967 (55.2\%) of the trees were Scots pine (Pinus sylvestris), 1381 (38.8\%) were Norway spruce (Picea abies), 188 (5.3\%) were of various unclassified deciduous species, and 26 (0.7\%) were contorta pine (Pinus contorta).

Finally, we leveraged data from the USA and Germany for a proper out-of-distribution (OOD) evaluation. 
The NEON data is from the National Ecological Observatory Network (NEON) site at the San Joaquin Experimental Range in California, USA. 
Most of the site consists of single-story canopy with herbaceous understory vegetation. 
The tree species in the area are live oak (Quercus agrifolia), blue oak (Quercus douglasii), and foothill pine (Pinus sabiniana). 
The BAMFORESTS datasets come from four different areas in and around Bamberg, Germany. 
The first dataset is from Hain, which is a forest-like park, located in the south-east of the city centre. 
The well-maintained forest of Hain consists mainly of deciduous trees with high variation in species due to its recreational nature.
The second dataset is collected from three different areas, one of which is in Stadtwald, located approximately 5 km to the south-east of Bamberg’s city centre, and the two others in Tretzendorf, located approximately 20 km west to the north-west of Bamberg.
These areas consist of managed forests with a mixture of coniferous and deciduous trees.

\subsection{Data}
\label{sec:main_data}

For studying the different tree crown segment estimation methods, we leveraged both remote sensing data and some form of tree crown reference from each area. 
The remote sensing data most importantly contained aerial imagery from each location, and from Espoonlahti we also utilised ALS-derived point cloud data. 
The used references varied from human-drawn image segments to semi-automatically collected harvester production record tree positions.
Figure~\ref{fig:dataset_features} visualises the distribution shift between all the used datasets.

\subsubsection{Espoonlahti}
\label{sec:espoonlahti_data}

From Espoonlahti, the data consisted of multispectral imagery and point clouds, with manually drawn pixel masks acting as the final evaluation reference.
The optical image data we used consisted of an orthophoto with a spatial resolution of 5~cm collected on 28 July 2023. The size of the complete orthophoto was 31\ 254 by 36\ 871 pixels, spanning an area of approximately 2~km$^2$. The multispectral imagery was captured from a helicopter with a MicaSense Altum-PT at an altitude of 120 m. MicaSense Altum-PT acquires five multispectral bands at 3.2 MP each, a 12.4 MP panchromatic band, and a thermal band. 
Agisoft Metashape Professional~\citep{agisoft}
was used for photogrammetric processing of the imagery using the "normalise band sensitivity" option, which applied relative radiometric normalisation based on camera calibration metadata (e.g., exposure, gain, black level, and vignetting correction), thereby improving inter-band and inter-image comparability without performing full reflectance calibration. 
A three-wavelength lidar system (HeliALS) installed on the helicopter was used to collect high-density point clouds from the study area simultaneously with multispectral image acquisition. A total of 66 ground control points extracted from HeliALS data were used during the bundle block adjustment step to improve alignment of the imagery. For further details of the HeliALS system, please refer to \citet{taher2026multispectral} and \citet{HYYPPA2026100142}.
For this study, only five Altum-PT multispectral bands were used. The bands were 475~nm (blue), 560~nm (green), 668~nm (red), 717~nm (red edge, RE), and 842~nm (near-infrared, NIR), with respective bandwidths of 32~nm, 27~nm, 16~nm, 12~nm, and 57~nm.

The multispectral ALS point clouds were acquired on 14 June 2016, during leaf-on conditions, using the Optech Titan sensor (Teledyne Optech, Ontario, Canada). TerraTec Oy (Helsinki, Finland) operated a fixed-wing data acquisition flight at an altitude of 700 m over the study area, generating three distinct point clouds corresponding to the 1550 nm (infrared), 1064 nm (NIR), and 532 nm (green) laser wavelength channels. 
The raw point clouds underwent initial radiometric and geometric calibration, which included intensity range-normalisation (further described in a work by~\citet{matikainen2017object}) and the removal of redundant flight-line overlaps to ensure spatial homogeneity. The resulting point clouds exhibited mean point densities of 11, 13, and 11 points/m$^2$ for channels 1, 2, and 3, respectively. All data were georeferenced to the ETRS-TM35FIN coordinate system, with vertical values referenced to the N2000 orthometric height system. 
A more detailed description of the lidar data acquisition and processing pipeline has been provided in the studies by~\citet{karila2019effect} and~\citet{taher2026multispectral}.

As the georeferencing of the multispectral orthophoto was done directly on the HeliALS data, using the 66 ground control points, the accuracy of the registration was mostly dependent on the sensors' spatial resolutions. The laser footprint on the ground surface for the HeliALS system was approximately 10~cm, and the resolution of the orthophoto was 5~cm. We estimate that the planar registration error for the orthophotos does not exceed 15~cm.
For the Optech Titan, the footprint in the used spectral bands was 30~cm~\citep{okhrimenko2019multi}, and therefore, we estimate that the planar registration error is below 35~cm. 

\begin{figure}[!ht]
    \centering
    \includegraphics[width=0.48\linewidth]{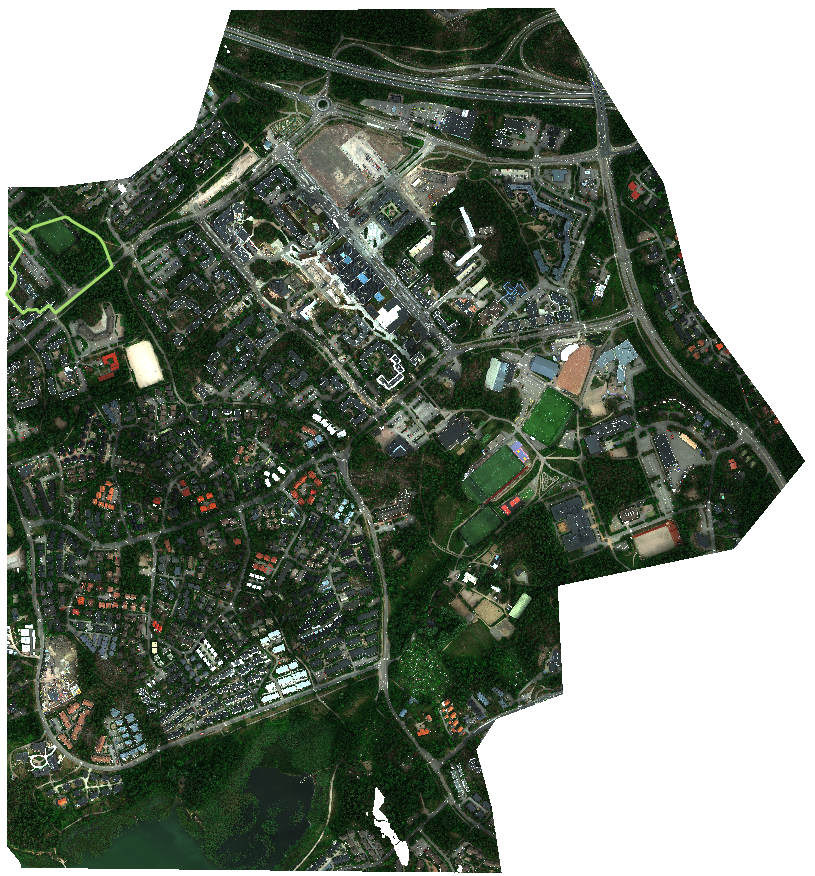}
    \includegraphics[width=0.48\linewidth]{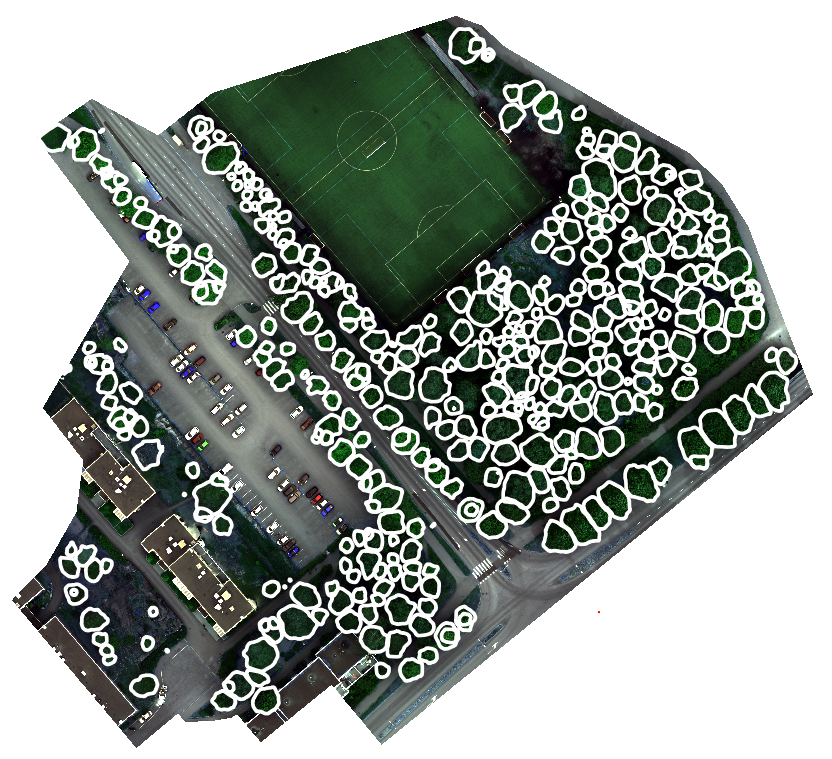}
    \caption{RGB visualisation of the orthophoto spanning the whole dataset region on the left, with the manually annotated test region delineated in light green and shown separately on the right. The manual test annotations are delineated in white.}
    \label{fig:orthos}
\end{figure}

We selected a single area of the orthophoto with a size of 4471 by 4893 pixels as a geographically separated test set. 
The trees in this area were manually annotated according to human perception of the same orthophoto. 
The temporal gap between the lidar data acquisition and multispectral image collection did not affect the test set, as the manual labels were based solely on the more recent imagery.
The orthophoto of the area and the separated test area are shown in Figure~\ref{fig:orthos}. The test area contained 362 manually segmented trees, and the rest of the orthophoto, used for training and validation, contained approximately 24\ 000 trees.
The test area was selected based on its various visual features as shown in Figures~\ref{fig:orthos} and~\ref{fig:labels}. 
The area contained roads, parking lots, buildings, and a football field alongside the different urban tree configurations, including an area where the resulting image patches represent a denser forested area with no visible features besides tree canopy.

For evaluation, the test ortho was split into predefined patches.
The default size of the patches was 1024 by 1024 pixels, resulting in 70 test images. 
The patch size, and thus the number of patches, was adjusted for some baseline methods to better match the data they were trained on.
This information is described in the corresponding baseline descriptions in Section~\ref{baselines}.

\subsubsection{Sundsvall}
\label{sec:sundsvall_data}

From Sundsvall, the image data were generated using the same hardware and preprocessing as for the data from Espoonlahti, described in Section~\ref{sec:espoonlahti_data}.
The Sundsvall imagery was captured on 18 August 2025, with harvester production records collected between 18 and 26 August 2025 acting as the reference data. 
The reference records contained approximate information on individual tree locations, as well as their size measured as the diameter at breast height (DBH). 

To evaluate how model performance varies with tree dominance and neighbourhood structure, we categorised the trees into four groups.
The four categories were dominant (higher than neighbouring trees), grouped (similar in height to neighbouring trees), semi-suppressed (with a neighbouring dominant tree more than 1.5 metres away), and suppressed (likely to be covered by a neighbouring dominant tree), following a study by~\citet{hakula2023individual}. 
As the reference data contained no tree heights, we estimated tree heights from the DBH information, using the Näslund function~\citep{b603dea5c4a74aa08d50ffedad328bbc}. 
The Näslund function coefficients were optimised using manually measured tree diameters and heights from Evo, Finland, which has a similar forest composition.
The tree height estimation process and the data from Evo are described in~\ref{app:naslund}.
The classification resulted in 1196 (33.6\%) dominant trees, 955 (26.8\%) grouped trees, 1009 (28.5\%) semi-suppressed trees, and 402 (11.3\%) suppressed trees.

\subsubsection{OOD data}
\label{sec:ood_data}

The NEON and BAMFOREST datasets, leveraged for the OOD evaluation, consisted of RGB imagery and manually drawn tree crown references.
The NEON dataset contained 18 RGB images and corresponding manually drawn tree crown instance delineations, defined as pixel masks. 
The imagery originates from the NEON Airborne Observation Platform as described by~\citet{weinstein2019individual}. 
The individual annotated images were sampled from a 1 by 1 kilometre orthomosaic with a spatial resolution of 0.1~m, which is part of the NEON high-resolution orthorectified camera imagery~\citep{https://doi.org/10.48443/xv9v-1r95}. 

The BAMFOREST datasets also consist of RGB imagery and manually drawn tree crown references. 
The imagery was collected using two drones, a small DJI Phantom 4  with a 1” CMOS 20MP sensor in the Hain area, and a larger Quantum Systems Trinity F90+ fixed-wing drone with a Sony RX1 RII in the other three locations. 
The raw imagery was collected in the summer of 2022 and used to create RGB orthomosaics with Agisoft Metashape (Version 1.8.4). 
The tree crown annotation process was done by an expert with access to additional data, and the final segments were confirmed with on-site checks.

\subsection{Pseudo-label generation}

The proposed method in this study relies on location-specific, automatically generated pseudo-labels. 
The pseudo-label generation process consists of two main steps: lidar-based coarse mask generation and SAM-based image segmentation mask enhancement.
Later in this study, the term coarse pseudo-labels corresponds to the initial lidar-based masks and enhanced pseudo-labels or simply pseudo-labels, to the final SAM-enhanced masks.
In this study, we generated these pseudo-labels for the Espoonlahti dataset, and the rest of the section refers to the pseudo-label generation process for that specific data.

\subsubsection{Coarse pseudo-labels}
\label{sec:methods:coarse_segs}

After the initial lidar data processing, individual tree crown labels were derived from the Optech Titan point clouds. 
Classification of returns into ground, vegetation, building, and noise classes was first performed using the LAStools software suite~\citep{lastools}.
A Digital Terrain Model (DTM) was then interpolated from the ground-classified points, enabling normalisation of vegetation returns to height above ground level.

Individual tree crown delineation was carried out using a variable‑window watershed segmentation approach by~\cite{kaartinen2012international}. A 0.5 m resolution CHM was generated from the combined first returns of Channels 1 and 2 (1550 nm and 1064 nm) to capture canopy structure in detail. The smoothing and local maxima filtering windows were scaled adaptively as a function of target canopy height. Specifically, the window diameters were set to 3, 5, 7, and 9 pixels for the height intervals of 0–7 m, 7–20 m, 20–30 m, and greater than 30 m, respectively. After smoothing the CHM with a Gaussian filter, treetops were identified using a local maximum filter. These treetop locations served as markers for a marker‑controlled watershed algorithm, which was applied to delineate individual crown boundaries. The resulting individual tree crown segments were subsequently used as pseudo‑labels in the downstream processing steps. 
The blockiness of the coarse pseudo-labels, shown in Figure~\ref{fig:labels}, was caused by the resolution difference between the generated CHM and that of the orthoimagery.
The resolution difference between the orthoimagery and CHM introduced a possible quantisation error with a maximum of half the CHM ground pixel size, meaning 25~cm.



As the point cloud was collected seven years earlier than the multispectral imagery, there were a large number of false positive segments in regions replaced by built environment. The false positive segments were filtered using a threshold of a red-RE index, which we denote as RREI, computed as
\begin{equation}
    \text{RREI} = (\text{RE}-\text{Red})/(\text{RE}+\text{Red}).
    \label{eq:ndvi}
\end{equation}
The index was computed from the multispectral orthophoto, and a threshold was set to 0.2, where segments with an average RREI less than that were discarded. 
The threshold was set based on qualitative estimates, where known, clear false positives were discarded while true positives were kept.
The index is equal to the numerator of the transformed red range vegetation index described by~\citet{blanco2020potential}, and it is similar to the normalised difference vegetation index (NDVI), with the sole difference that the NIR channel of NDVI has been replaced with that of the RE. For both indices, 0.2 is used as the typical threshold for vegetation~\citep{bhandari2012feature,hashim2019urban}.
A histogram of the RREI distribution of the segments, shown in Figure~\ref{fig:ndvi_hist} of~\ref{app:figures}, also highlights that most segments fall either clearly over or clearly under the set threshold.  


\subsubsection{SAM 2 enhancement}
\label{sec:methods:pseudo_segs}


Due to the resulting CHM resolution of the previous process, the initial tree crown segments were coarse compared to the resolution of the available aerial imagery. 
To generate more precise annotations, we leveraged SAM 2, a model where the inputs are an image and queries, such as points or bounding boxes, and the outputs are segments which correspond to said queries. To generate masks of the tree crowns, we used bounding boxes of the coarse labels (the smallest rectangle containing all coarse label pixels) as the query inputs for SAM 2. 
We opted for the bounding box prompting strategy as the segments we obtained using simple centre point prompts yielded visibly worse results in early tests and more complex multi-prompt strategies would have required additional assumptions about the tree crowns.

\begin{figure}
    \includegraphics[width=\linewidth]{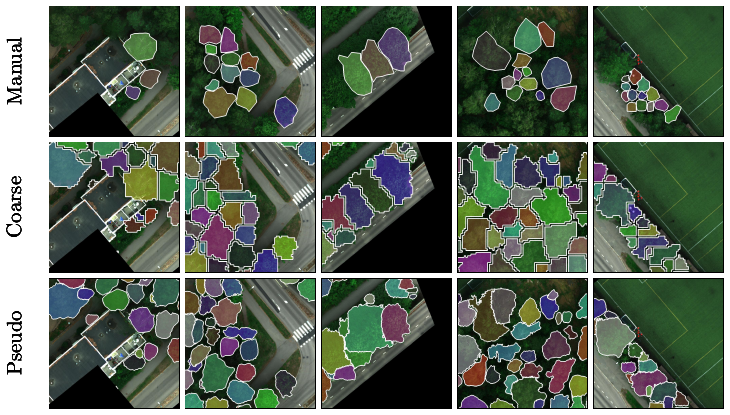}
    \caption{Visualisation of the different labels on the test set. The pseudo-randomly selected colours are used only to distinguish separate instance segments from each other.}
    \label{fig:labels}
\end{figure}

To avoid segmenting only partly visible trees, the orthophoto was sampled in patches of 1024 by 1024 pixels using a stride of 512 pixels. This meant that all trees could be captured by only pseudo-labelling the trees, whose centroids were found in the 512 by 512 pixel square in the middle of each patch. This specific configuration was chosen, as from the distribution of tree crown sizes, it was found that only a tiny fraction of the trees had large enough crowns that they could extend from the middle area to the outside of the image.
During the pseudo-labelling process, SAM 2 first processed the image, then the corresponding bounding boxes were sampled iteratively to prompt SAM 2 to generate each separate mask.
The SAM 2-generated pseudo-labels are visualised on five test set samples in Figure~\ref{fig:labels} alongside the manual and coarse segments. 

\subsection{Proposed pseudo-supervised model}
\label{sec:training}

\begin{figure}[!ht]
    \centering
    \includegraphics[width=0.8\linewidth]{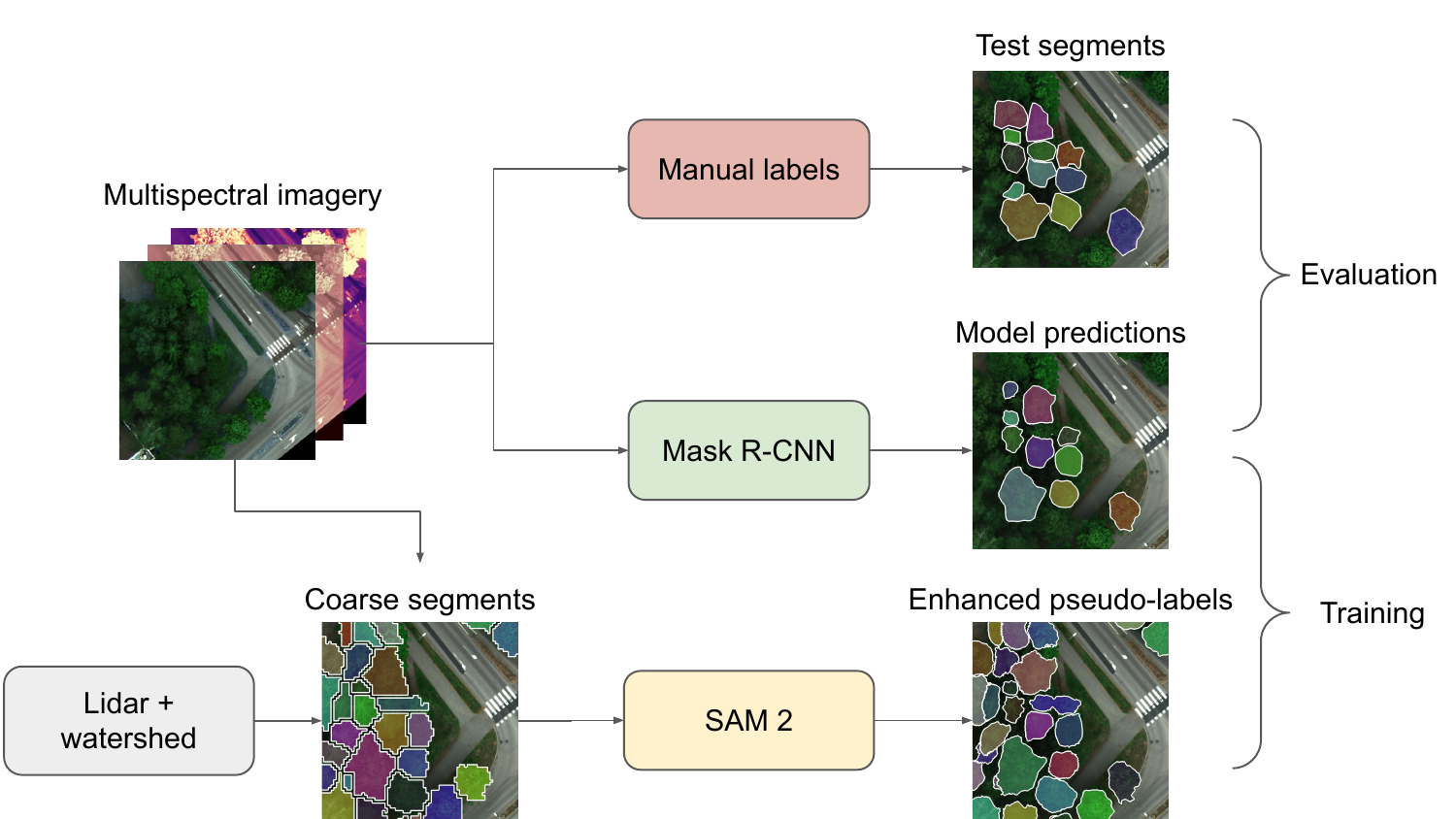}
    \caption{Overview of the proposed data fusion method for the pseudo-supervised model training.}
    \label{fig:overview}
\end{figure}

We used the CHM and SAM 2-derived pseudo-labels to train downstream models which directly predict individual tree crown masks from RGB or multispectral images. An overview of the model training and evaluation scheme is shown in Figure~\ref{fig:overview}. 
As our model architecture, we chose an improved version of the Mask R-CNN~\citep{he2017mask,li2021benchmarking} with an ImageNet~\citep{deng2009imagenet} pretrained ResNet 50 FPN backbone~\citep{he2016deep,lin2017feature}. 
The specific configuration was chosen due to its prominence in both general and tree crown instance segmentation literature~\citep{minaee2021image,ball2023accurate}. 
Choosing a widely tested model for our implementation assured that any performance improvements could mostly be attributed to the training data instead of the model architecture when comparing the models on our evaluation datasets as well as to the metrics reported in the literature. 
Possible future improvements and our recommendations for obtaining the best-performing model for the task in general are highlighted in Section~\ref{sec:discussion}.

For multispectral variants using RGB and RE or NIR bands, we adapted the model by repeating the parameters of the first image layers. This allowed us to evaluate the effects of different input channels on the capability of the model. However, it has to be noted that all pseudo-labels were generated using only RGB imagery.  
The models were optimised based on the simple sum loss function $L = L_{cls} + L_{box} + L_{mask}$, where $L_{cls}$ is a classification loss, $L_{box}$ is a box prediction loss, and $L_{mask}$ is a mask loss. The loss is described in detail in the original work introducing the Mask R-CNN architecture~\citep{he2017mask}. 
The models were trained using the Adam optimiser~\citep{kingma2014adam} with decoupled weight decay~\citep{hanson1988comparing,loshchilov2017decoupled}. 
Both the training and validation loss were computed between the automatically generated coarse labels and the SAM 2-enhanced pseudo-labels, corresponding to whichever data were used for training. 
During the training process, a pre-sampled set of image patches, sized 1024 by 1024 pixels, was used for both training and validation. The validation set was sampled with a pseudo-random algorithm, and the same validation split was used for all model variations. The validation set contained 209 (19\%) images, and the training set 886 (81\%). The training images were augmented with horizontal and vertical flipping, random rotations [-180°, 180°], and random IoU crops, as introduced in the work of~\citet{liu2016ssd}. The random IoU crops varied between scales of 0.3 to 1.0, and aspect ratios of 0.5 to 2.0. Each augmentation was performed independently and sampled uniformly in the corresponding range.

The validation set was used to choose the best model obtained during a training process of 400 epochs, and to optimise the training and inference hyperparameters through trial and error. The final test set evaluated models were trained with a learning rate of 1e-6, batch size of 8, and weight decay of 1e-2. For inference, we used a mask threshold of 0.5, a score threshold of 0.3, and a non-maximum suppression (NMS) IoU of 0.3. NMS removes overlapping output masks based on the set threshold, keeping the mask with a higher score.

\subsection{Baseline image-based models}
\label{baselines}

We compared the pseudo-supervised Mask R-CNN models to various baselines. As our method could be deployed without the requirement of locally collected, manually annotated training data, we dismissed alternatives which required such data. Thus, the baseline methods represent alternative solutions to obtain tree crown segment predictions without leveraging local, manually annotated data for either training or fine-tuning. 
We also considered both the coarse and SAM 2-enhanced masks as alternative tree crown segmentation methods, as they could be deployed directly in locations with ALS data available, such as the Espoonlahti test area of this study. We only considered methods for which the source code was publicly available. For testing, we used the same inference parameters as for our proposed pseudo-supervised models, unless specified in the following corresponding paragraphs.

\paragraph{\textbf{Detectree2}} A family of fully supervised Mask R-CNN models trained using other available tree crown instance segmentation data, called Detectree2, is one of the most prominent models used for the task~\citep{ball2023accurate}. At the time of this study, seven pretrained Detectree2 models were available, and we evaluated the performance of each. Each of the models used the same Mask R-CNN architecture with a ResNet 101 FPN backbone, and they differed from each other in their training data compositions and training strategies.

The pretrained model checkpoints were available on Zenodo~\citep{ball_2025_15863800}. Following the authors' naming convention, the models are called \textit{withParacouUAV}, a model trained with UAV and aeroplane captured imagery from tropical areas, \textit{randresize\_full}, a model trained with the same data but also using a random resize augment, \textit{urban\_trees\_Cambridge}, a model trained with urban data from Cambridge, UK, \textit{base}, a model trained with aeroplane captured tropical data and introduced as the \textit{base} model in an earlier work by~\citet{Ball2024.06.24.600405}, \textit{05dates}, a copy of the \textit{base} model further tuned with $\sim$5~cm resolution UAV-captured RGB imagery from five different dates and referred to as the \textit{5 date} model in the same prior work by Ball et al., \textit{flexi}, a model trained with data from both urban and forested areas, and \textit{tropical\_closed\_canopy}, a later version of a model focused on tropical forest areas. As all the models were trained with image patches of approximately 1000 by 1000 pixels with spatial resolutions varying from approximately 5 to 10~cm, we deployed the models directly on the same 1024 by 1024 pixel patches as our proposed methods.

\paragraph{\textbf{Supervised U-Net}} A publicly available, fully supervised tree crown instance segmentation model has also been developed based on the U-Net architecture~\citep{freudenberg2022individual}. The model has been tested with $\sim$5cm resolution aerial imagery, and we deployed the pretrained model on our data as is. Besides the RGB image channels used by the other baseline models, the U-Net also uses the NIR channel with the red colour channel to construct an NDVI image, where each pixel value is computed as
\begin{equation}
    \text{NDVI} = (\text{NIR}-\text{Red})/(\text{NIR}+\text{Red}).
    \label{eq:ndvi2}
\end{equation}
The NDVI image is then used as an additional input channel by the U-Net model.
At the time of this study, model weights were only available for a model that uses the NDVI, and the corresponding study also did not report results from any RGB-only model. 
The model has been developed for inference on image patches of size 224 by 224 pixels. As such, we investigated the accuracy of a few options with different image patching and resizing configurations. In the results, we report the best obtained metrics with the configuration of patching the ortho into images of size 512 by 512 pixels and resizing them to 224 by 224 pixels before passing them to the neural network. 

\paragraph{\textbf{Grounded SAM (2)}} A fusion of Grounded DINO~\citep{liu2023grounding} and Segment anything model~\citep{kirillov2023segany}, and the latter version leveraging SAM 2~\citep{ravi2024sam2segmentimages}. The model leverages zero-shot bounding boxes from the Grounded DINO model to prompt SAM (2)~\citep{ren2024grounded}. The method has been applied in the context of tree crown segmentation to aid in the construction of the FMARS dataset~\citep{arnaudo2024fmars}. The authors of the FMARS dataset stated that the prompt "bushes" worked the best for detecting trees in aerial images. We tested this alongside a few other prompts and arrived at the same conclusion.

\paragraph{\textbf{DeepForest + SAM 2}} DeepForest is a bounding box detection model for tree crowns. It has been trained with various tree crown data, promising at least some generalisability across datasets~\citep{weinstein2020deepforest}. To leverage the model for segmentation, we combined it with SAM 2, which was used by prompting it with the bounding boxes generated by DeepForest.

\paragraph{\textbf{SAM 3}} The latest version, 3, of the Segment Anything Model comes alongside a native feature for text prompting, meaning that it could be directly deployed on images without fine-tuning or leveraging external object detection models~\citep{carion2025sam3segmentconcepts}. 
The text prompt we used for SAM 3 was simply "tree", whereas the prompt "bushes", which worked for Grounded SAM, yielded worse results.

\subsection{Evaluation}
\label{sec:methods:eval}

To evaluate the different tree crown segmentation methods, we leveraged two types of evaluation metrics depending on the type of available reference. 
Most results presented in this study are standard image segmentation metrics: mIoU, F1 score, precision, and recall.
For the precision, recall, and F1 computations, a predicted tree crown was considered a true positive if it had an IoU of at least 0.5 with a ground-truth crown.
In addition, on the Espoonlahti dataset, to account for the small size of our test set, we estimated the 95\% confidence intervals of our results using test set bootstrapping. The bootstrapping was performed at the individual tree level with 1000 resampled sets.
The confidence intervals provide a more representative estimate of how well the method holds for other data, discounting significant domain shifts.

To compute evaluation metrics for the Sundsvall data, we leveraged the tree position-based metrics.
The evaluation process required first generating tree crown predictions for the whole referenced area and extracting tree position estimates based on the predicted segment centre points.
These predicted tree positions then had to be matched to reference tree positions. 
The matching process was done in two ways, one-to-one and one-to-many, by minimising the mean Euclidean distance in the two ground plane dimensions between the predicted and reference tree position sets. 
One-to-one matching means that each predicted tree crown was matched to a single reference tree position and vice versa. 
One-to-many was only restrictive one way, such that a single reference could not be matched to multiple predicted trees, but a single prediction could be matched to multiple references. 
The latter approach was used to provide evaluation which is more forgiving of the fact that many of the smaller trees could be completely covered by larger tree crowns, rendering them invisible from the aerial perspective. 
In addition, metrics were computed for the different tree visibility categories, described in Section~\ref{sec:sundsvall_data}.
The metrics computed from the matched sets were mean matched tree distance, F1-score, precision, and recall. 
For both matching strategies, a threshold of 1.5 metres was used as the distance threshold.

\section{Experiments and results}

We conducted experiments to compare the different tree crown instance segmentation methods and to ablate our proposed model using our main dataset from Espoonlahti. 
We performed model transferability analysis on the other datasets, and finally, tested the computational efficiency of the most relevant models.

\subsection{Method comparison}

The quantitative results of the method comparison are presented in Table~\ref{tab:main_results}. 
The comparison includes non-local image-based baselines, directly generated CHM-based and SAM 2-enhanced masks, and our pseudo-supervised models trained using the locally generated pseudo-labels.
The RGB-only and best-performing RGB+NIR variants of our proposed models are highlighted at the bottom, with the baseline methods above.
The results show that the proposed method outperforms all evaluated alternatives on the Espoonlahti dataset. 
The best-performing purely image-based baseline, the supervised U-Net, achieved a higher mIoU than the directly evaluated CHM-derived masks but remained below the proposed methods. Other notable results among the baseline methods were those of SAM 3 and Detectree2. While the mIoU and F1 scores were clearly worse than those obtained with our final models, the high precision of the models indicated relatively few false positives. However, this was in part caused by the models being overly cautious, only finding a few of the visible tree crowns in the image, as highlighted by the low recall values. 

\begin{table}[!ht]
    \centering
    \caption{Comparison between our proposed methods (bottom) and the alternative image-based baselines (top), with the lidar-based results in the middle. The best results are bolded and the second best underlined. *Detectree2 results correspond to the best results obtained with the \textit{Flexi} checkpoint. Full results from all seven checkpoints are shown in Table~\ref{tab:detectree2} of~\ref{app:detectree2}.}
    \begin{tabular}{lrrrrr}
        \toprule
        \textbf{Method} & \textbf{F1} & \textbf{Precision} & \textbf{Recall} & \textbf{mIoU} & \textbf{mIoU 95\% CI} \\
        \midrule
        Detectree2* & 0.556 & \textbf{0.911} & 0.400 & 0.324 & [0.281, 0.364] \\
        U-Net & 0.303 & 0.205 & 0.583 & 0.524 & [0.501, 0.547] \\
        Grounded SAM & 0.166 & 0.430 & 0.103 & 0.094 & [0.070, 0.121] \\
        DeepForest & 0.501 & 0.544 & 0.464 & 0.396 & [0.363, 0.428] \\
        SAM 3 & 0.558 & \underline{0.822} & 0.422 & 0.352 & [0.311, 0.397] \\
        \midrule
        Coarse masks & 0.571 & 0.550 & 0.594 & 0.477 & [0.447, 0.505] \\
        Pseudo masks & 0.597 & 0.584 & 0.611 & 0.513 & [0.478, 0.545] \\
        \midrule
        RGB & \underline{0.758} & 0.783 & \underline{0.733} & \underline{0.599} & [0.566, 0.632] \\
        \textbf{RGB NIR} & \textbf{0.778} & 0.783 & \textbf{0.772} & \textbf{0.621} & [0.587, 0.652] \\
        \bottomrule
    \end{tabular}
    \label{tab:main_results}
\end{table}

\begin{figure}[!ht]
    \includegraphics[width=0.95\linewidth]{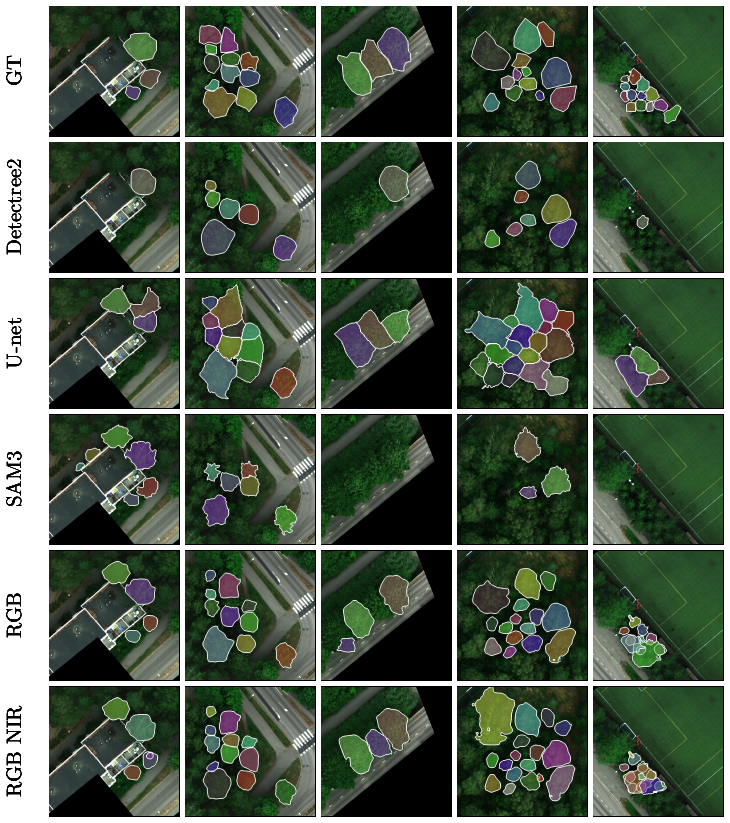}
    \caption{Visualisation of the outputs of selected models. The Detectree2 results are from the best-performing \textit{Flexi} checkpoint. The pseudo-randomly selected colours are used only to distinguish separate instance segments from each other. Each column represents a single test image with outputs from different models, with rows representing different models as indicated on the left-hand side.}
    \label{fig:results_vis}
\end{figure}

Qualitative results from selected models are shown in Figure~\ref{fig:results_vis}, while the outputs of baseline methods not shown in Figure~\ref{fig:results_vis} are presented in Figure~\ref{fig:main_visuals_app} of~\ref{app:figures}. The visualised test samples were chosen to highlight the variety of scenarios present in the data, purely based on the input and ground truth masks. The frames are cropped to 3/4 of the original width and height of 1024 pixels for the visualisation, as only the masks in the middle 512 by 512 pixel area were evaluated and thus also visualised for each patch. 
The qualitative results support our quantitative results, as the proposed models generally reproduce the number and extent of the reference tree crowns more closely than the baseline methods, resulting in more balanced outputs. 
The slight improvement of the RGB+NIR model over the RGB-only model can also be seen in the same figure. 
Detectree2 and SAM 3 models appear too conservative in their predictions, while the supervised U-Net model produces more false positives and overly large segments. 
These observations are consistent with their measured precision and recall.

\subsection{Ablation}

\begin{figure}[!ht]
    \centering
    \includegraphics[width=0.8\linewidth]{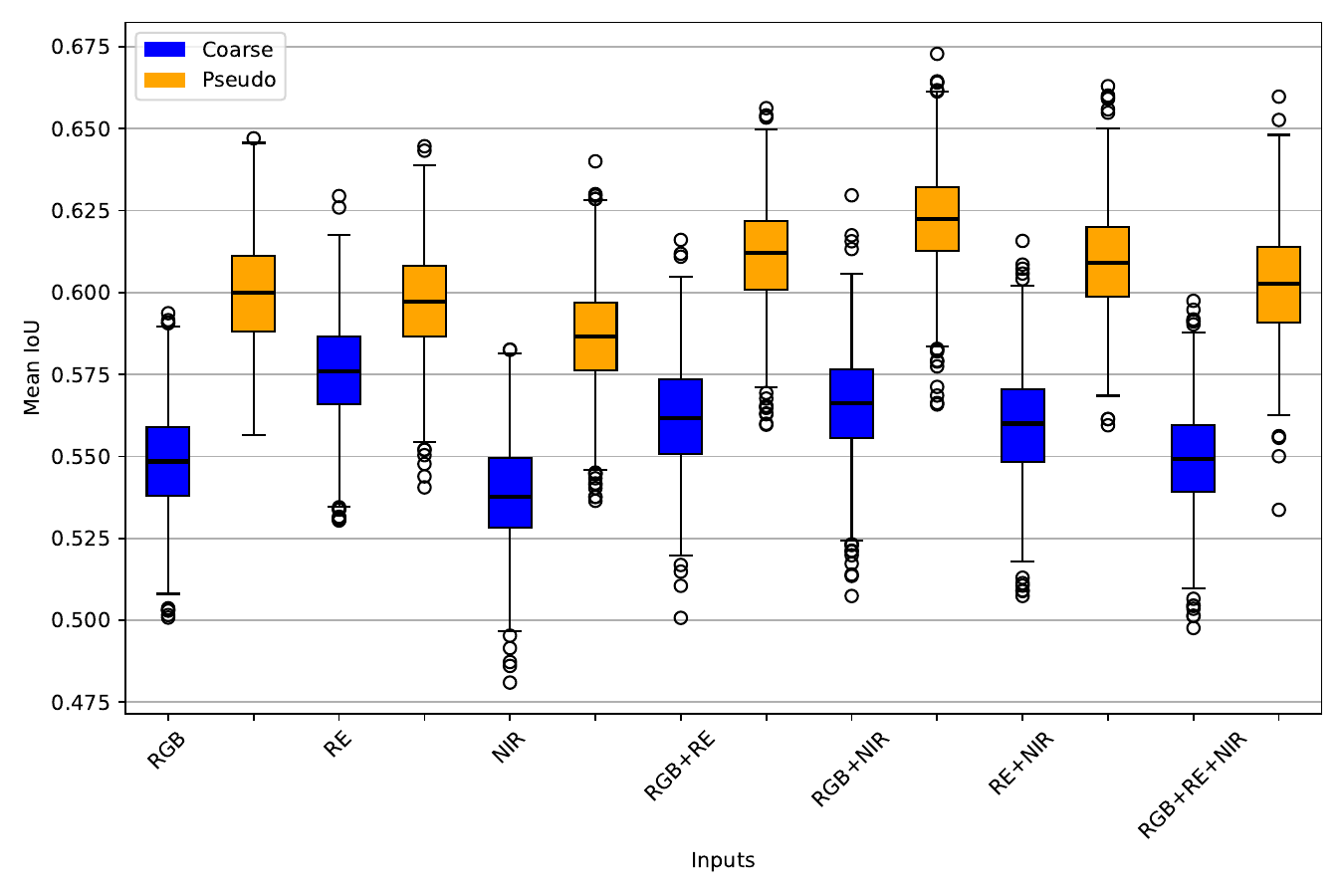}
    \caption{Standard quartile box plot showing the prediction quality (mIoU) between different input modalities and coarse vs. pseudo-supervision.}
    \label{fig:supervision_boxplot}
\end{figure}

To verify the usefulness of the SAM 2 enhancement step in the pseudo-label generation process, we evaluated models with different channel configurations trained with both the coarse and enhanced pseudo-labels. 
The results are shown in Table~\ref{tab:ablation:input}, with the mIoU test set bootstrapping distributions visualised in Figure~\ref{fig:supervision_boxplot}. 
Table~\ref{tab:ablation:input} shows the standard metrics used for all of the evaluated methods.
The validation losses were computed from a consistent pseudo-random subset of training data, meaning either coarse or enhanced lidar-based segments, respectively. 
As shown, the SAM 2 label enhancement improves the downstream model mIoU across all tested input configurations. 
The best model mIoU (RGB+NIR) increased by 9.9\% from 0.565 to 0.621, and the average mIoU across tested models increased by 8.4\% from 0.557 to 0.604. 

\begin{table}[!ht]
    \centering
    \caption{Ablation results of the different trained Mask-RCNN models. The confidence interval \textbf{95\% CI} is that of the mIoU. For the validation loss, lower is better. For all other metrics, higher is better. The best results are bolded and the second best underlined. The validation losses were computed using the respective coarse or pseudo-labels from a pseudo-random validation subset.}
    \begin{tabular}{llrrrrrr}
        \toprule
        \textbf{Supervision} & \textbf{Input} & \textbf{Val. loss} & \textbf{F1} & \textbf{Precision} & \textbf{Recall} & \textbf{mIoU} & \textbf{95\% CI} \\
        \midrule
        Coarse & RGB & 1.912 & 0.743 & 0.784 & 0.706 & 0.548 & [0.515, 0.579] \\
        Coarse & RE & 1.964 & \underline{0.775} & \textbf{0.805} & 0.747 & 0.576 & [0.541, 0.607] \\
        Coarse & NIR & 1.965 & 0.739 & 0.777 & 0.706 & 0.539 & [0.505, 0.568] \\
        Coarse & RGB+RE & 1.903 & 0.740 & 0.766 & 0.717 & 0.562 & [0.528, 0.591] \\
        Coarse & RGB+NIR & 1.913 & 0.746 & 0.762 & 0.731 & 0.565 & [0.532, 0.595] \\
        Coarse & RE+NIR & 1.925 & 0.760 & 0.792 & 0.731 & 0.559 & [0.525, 0.590] \\
        Coarse & RGB+RE+NIR & 1.923 & 0.748 & 0.786 & 0.714 & 0.549 & [0.515, 0.579] \\
        
        Pseudo & RGB & \underline{1.732} & 0.758 & 0.783 & 0.733 & 0.599 & [0.566, 0.632] \\
        Pseudo & RE & 1.776 & 0.767 & 0.776 & \underline{0.761} & 0.597 & [0.565, 0.631] \\
        Pseudo & NIR & 1.813 & 0.767 & 0.785 & 0.750 & 0.586 & [0.554, 0.617] \\
        Pseudo & RGB+RE & \textbf{1.723} & 0.774 & \underline{0.793} & 0.756 & 0.610 & [0.577, 0.643] \\
        Pseudo & RGB+NIR & \underline{1.732} & \textbf{0.778} & 0.783 & \textbf{0.772} & \textbf{0.621} & [0.587, 0.652] \\
        Pseudo & RE+NIR & 1.753 & 0.772 & 0.787 & 0.758 & \underline{0.610} & [0.576, 0.640] \\
        Pseudo & RGB+RE+NIR & 1.738 & 0.761 & 0.771 & 0.750 & 0.602 & [0.567, 0.635] \\
        \bottomrule
    \end{tabular}
    \label{tab:ablation:input}
\end{table}

\begin{figure}
\centering
    \includegraphics[width=0.9\linewidth]{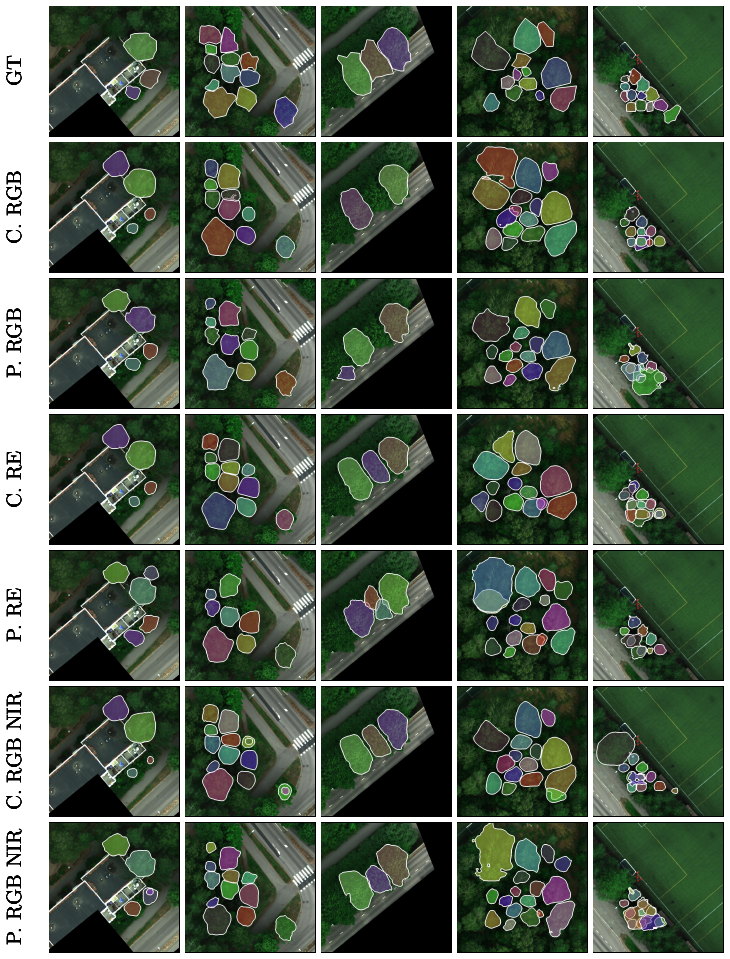}
    \caption{Visualisation of the outputs from models using different input modalities and training labels. C. is short for coarse and P. for pseudo-labels, respectively. The pseudo-randomly selected colours are used only to distinguish separate instance segments from each other. Each column represents a single test image with outputs from different models, with rows representing different input channel configurations as indicated on the left-hand side.}
    \label{fig:qual_ablation}
\end{figure}

The same experiment also gave insight into the value of different input channels.
However, the effect of different input channel configurations is less clear. 
Notably, the model using the sole RE image input behaves slightly differently from the other models. 
Specifically, its performance was more robust than the other models' when trained using the coarse segments, achieving the highest mIoU among the coarse-supervised models. 
Example frames of different input channels are shown in Figure~\ref{fig:input_channel_comp} of~\ref{app:figures}.
The qualitative results of different input modalities and training label configurations can be observed from Figure~\ref{fig:qual_ablation} and Figure~\ref{fig:abl_rest} of~\ref{app:figures}.

Finally, we evaluated how the inference time image patching process can affect the proposed models' performance. 
The experiment showed that on average, the mIoU of the models decreased by 3.33\% when evaluated across the whole image rather than limiting it to the middle quarter.
The metrics resulting from an evaluation with full image stride are shown in Table~\ref{tab:results:full_stride} of~\ref{app:edge_eval}.
Only the model using the RE channel input was the sole outlier in this regard, and we presume that it is similar to the behaviour observed on the coarse segment training, where surprisingly the solely RE-using model outperformed the others in some metrics. 

\subsection{Cross-dataset performance}

To assess the proposed models' transferability to data from different settings, we evaluated the models using the four other datasets. 
The proposed RGB+NIR model was evaluated on the Sundsvall dataset in Sweden.
The metrics show that the model can detect trees in an environment different from the training data while using the same sensor and preprocessing workflow. 
The overall performance was slightly lower than on the Espoonlahti data, which could be expected given the large difference in the environmental composition. 
The dominance class-specific results show a clear trend towards better detection of dominant than suppressed trees. 
Recall is the most informative metric for the class-specific results because the predicted segments were not assigned to any dominance classes. 
Consequently, we calculated precision and F1 using all 2881 predicted segments. 
This caused relatively low class-specific precision and F1 values, which are not directly comparable to the overall metrics.

\begin{table}[!ht]
    \centering
    \caption{The results from the Sundsvall test data.}
    \begin{tabular}{llrrrr}
        \toprule
        \textbf{Class} & \textbf{Matching} & $\boldsymbol\mu$\textbf{(dist.) (m) }& \textbf{Precision} & \textbf{Recall} & \textbf{F1} \\
        \midrule
        Total & 1-to-1 & 2.119 & 0.572 & 0.463 & 0.512 \\
        Total & 1-to-many & 1.572 & 0.668 & 0.540 & 0.598 \\
        grouped & 1-to-1 & 1.610 & 0.194 & 0.585 & 0.291 \\
        grouped & 1-to-many & 4.414 & 0.210 & 0.635 & 0.316 \\
        semi-suppressed & 1-to-1 & 1.692 & 0.183 & 0.521 & 0.270 \\
        semi-suppressed & 1-to-many & 3.778 & 0.202 & 0.578 & 0.300 \\
        Suppressed & 1-to-1 & 1.864 & 0.063 & 0.450 & 0.110 \\
        Suppressed & 1-to-many & 7.575 & 0.069 & 0.498 & 0.122 \\
        Dominant & 1-to-1 & 1.453 & 0.289 & 0.696 & 0.409 \\
        Dominant & 1-to-many & 2.963 & 0.300 & 0.722 & 0.424 \\
        \bottomrule
    \end{tabular}
    \label{tab:sundsvall_intext}
\end{table}

In contrast, the tests showed that the proposed RGB-model did not generalise to the three public datasets. 
When tested on the manually annotated subset of the NEON tree segmentation dataset~\citep{locke_2025_15420565}, the model detected only two of the 148 labelled trees in the imagery. 
Similarly, on the two separate BAMFOREST test sets~\citep{troles2024bamforests}, the model obtained mIoUs of 0.124 and 0.038.

\subsection{Computational efficiency}

While the main intention of our method was not to create a smaller, more efficient model, the process distils the useful information from the larger SAM 2 model into the smaller Mask R-CNN model. 
We evaluated the computational efficiency of the better-performing baseline models, the supervised U-Net and SAM 3, which also highlight the range of possible model sizes for the task, with the former representing the lighter available models and the latter the heavier.  
The results are shown in Table~\ref{tab:comp_eff}.
The proposed models' computational requirements fall between the reference points but far closer to the lighter end.
The commonly used Detectree2 model has approximately the same computational efficiency metrics as our proposed model, due to sharing a common model architecture with a slightly heavier image encoder.

\begin{table}[!ht]
    \centering
    \caption{Number of parameters of different tested models and their computational complexity in terms of inference speed and number of parameters. Inference times were evaluated on an NVIDIA RTX A4000 Laptop GPU.}
    \begin{tabular}{lrr}
        \toprule
       \textbf{Model} & \textbf{Inference time (ms)} & \textbf{Number of parameters (M)} \\
       \midrule
        U-Net & 70 & 31 \\
        Mask R-CNN (Ours) & 90 & 46 \\
        SAM 3 & 290 & 891 \\ 
        \bottomrule
    \end{tabular}
    \label{tab:comp_eff}
\end{table}

\section{Discussion}
\label{sec:discussion}

The experiments showed that CHM-generated annotations can be used to train image-based tree crown segmentation models and that SAM 2-based label enhancement improves both the pseudo-label quality and downstream segmentation performance.
Notably, the models trained with automatically generated and SAM 2-enhanced pseudo-labels outperformed the pseudo-labels themselves when evaluated against the manual reference annotations, suggesting that the downstream models can learn useful representations despite imperfections in the training labels.

\subsection{Main findings}

The proposed models outperformed the baseline image-based models on the data from Southern Finland.
The relatively poor performance of several baseline methods may be related to the domain shift between their training data and the Espoonlahti data, and the domain shift is also reflected in the proposed methods' performance on the public datasets.
The contrast between the Sundsvall and public-dataset results suggests that transferability was substantially better when sensor characteristics and preprocessing remained consistent, whereas larger shifts in image characteristics and spatial resolution remained challenging.
Heavier data augmentations could likely alleviate the problems caused by data distribution shifts, but these were yet to be explored further.
The obtained metrics for the proposed pseudo-supervised models fell into the range that has been observed in previous studies, where models have been trained with manually drawn training data in different locations.
Namely, our best obtained mIoU of 0.621 falls clearly within the range of 0.49 to 0.94 seen in prior work~\citep{wolk2024review}.

Intuitively, the result of the pseudo-supervised models outperforming the zero-shot baselines was not surprising, as the models leveraged training data from the same location. 
However, the key benefit of the proposed approach is that this local supervision can be generated automatically from existing ALS point clouds and imagery without manual crown delineation.
In the comparison between the baseline methods, it is notable that the supervised U-Net~\citep{freudenberg2022individual} had clearly the best performance.
The SAM 3~\citep{carion2025sam3segmentconcepts} and Detectree2~\citep{ball2023accurate} models demonstrated high precision at the cost of overly cautious predictions, meaning they completely missed many of the reference trees. 
The U-Net model's success could possibly be attributed to the fact that it was the only model that also leveraged the available NIR image through the generated NDVI image, while the other models had only been trained on RGB images. 
This aligns with our result where the RGB+NIR model achieved the best mIoU among the proposed models, suggesting that the NIR channel may present significant benefits for the task.
It could also be an indication that the distribution shift between datasets is less evident in the NIR channel, or that the specific training data were closer to the distribution of our test data.
Unfortunately, we did not have access to the training data or models trained solely on RGB data from the U-Net study to test these hypotheses.

The ablation provided concrete evidence of the usefulness of the SAM 2 label enhancement, but the results regarding different input channel configurations were less conclusive. 
To confidently determine the optimal input channel configuration for the task, more experiments with more varied data should be carried out.
Notably, the model using the sole RE image input behaves slightly differently from the other models. 
Specifically, its performance was more robust than the other models' when trained using the coarse segments, achieving the highest mIoU among the coarse-supervised models. 
We suspect that this might have been caused by the tree crowns in the RE image occasionally being less clear than in the other channels, causing the model to be less prone to overfitting to undesired features, such as the blockiness present in the coarse labels. 
This is supported by the fact that the validation loss of the model was worse than that of most other models, highlighting the disagreement between the coarse validation labels and the manually drawn test labels.
The visualisation of the different channels in Figure~\ref{fig:input_channel_comp} in~\ref{app:figures} also highlights these differences.
Methods to use the RE channel should be investigated further, but the behaviour might also be dataset-specific and related to, for example, channel normalisation.
An additional practical benefit of the proposed approach is its potential to exploit previously acquired lidar datasets for generating training annotations. In this study, the ALS data were acquired seven years before the optical imagery, yet the derived crown segments could still be used to train effective image-based segmentation models. The SAM 2-based enhancement, together with vegetation-based filtering, helped improve the usability of these older data for pseudo-label generation. This is particularly relevant in regions where previously acquired ALS datasets are available but up-to-date tree crown annotations are lacking.

\subsection{Scope and limitations}

The intention of the study is to highlight an efficient way of obtaining domain-specific data to tackle a distribution shift that is well present in most data available for segmenting individual tree crowns from orthoimagery. 
Most of our experiments were limited to a single region due to the lidar-based CHM and visual data availability. 
As such, our study mostly demonstrates that the data generation method can be used to obtain location-specific models efficiently. In addition, the manually annotated Espoonlahti test set was relatively small and represented only a single geographic area, limiting the environmental diversity of the primary evaluation.
We still believe the results are representative of the methods' real performance in such scenarios, as the area contains various visual features and types of trees, with regard to species, sizes, and shapes. 

While our RGB+NIR model demonstrated some generalisation capability on the Sundsvall dataset, we lacked extensive data from more varied sources. Because the Sundsvall data were acquired using the same sensor and preprocessing workflow as the training data, this experiment represents only a partial domain shift and does not demonstrate cross-sensor generalisation. The Sundsvall evaluation also relied on approximate tree locations and dominance classes derived from DBH-based height estimates, introducing additional uncertainty into the class-specific analyses.
In addition, the observed differences between different input modalities were relatively small, and slightly better segmentation metrics compared to human annotations do not necessarily translate directly into better real-world performance~\citep{khan2025validation}.

The training data also presented some additional challenges caused by the seven-year temporal gap between the lidar and image data collection dates. 
The RREI-based filtering could not entirely remove the problems caused by the temporal gap either, as it does not take into account tree growth. 
Thus, the temporal mismatch likely introduced additional noise into the training labels, but the proposed models still achieved strong performance.
However, part of the positive effect of the SAM 2 label enhancement could likely be attributed to the new segments getting automatically adjusted to the more recent imagery compared to the coarse segments derived entirely from the older lidar data. 
The coarse and enhanced lidar-based segments were likely the most affected by the temporal gap, meaning that their performance might have been slightly underestimated.
Some of the effects of the SAM 2 enhancement might also be achievable with specific heuristics which could estimate tree growth, smooth the labels to better match the resolution of the optical imagery, and better filter any false positives. 
However, this would require a significant amount of manual engineering compared to the simple alternative of leveraging publicly available foundation models. 

We focused our ablation on the use of SAM 2 for the training label generation and on the use of different input channels by the image-based model, as the other components of our methodology have already been studied extensively in prior literature. 
Specifically, both watershed segment generation and Mask R-CNN have been used extensively for unsupervised lidar-based segmentation~\citep{kaartinen2012international,zhen2016trends} and supervised image-based segmentation, respectively~\citep{minaee2021image}. 
In the ablation, we also identified two weaknesses of our proposed models related to inference:

\paragraph{\textbf{Overlapping predictions}} 

The main weakness of the proposed methods, which could be observed on the used test samples, was the tendency to produce outputs in which some masks are entirely contained by larger predicted masks. For masks which were of similar size, this was not a problem, as they are removed by the NMS post-processing step, but in scenarios where one of the masks was significantly larger than the other, the NMS could not capture the overlap, as the filtering was done based on a simple intersection-over-union threshold. However, for practical inference, another filtering step could be included based on the intersection and the area of a single mask. The problem also highlights the challenges in the Espoonlahti dataset caused by the large differences in the size of the presented trees. 

\paragraph{\textbf{Predictions near image edges}}

As the models were trained on labels which were only present in the middle part of the image, the predictions deteriorated closer to the edges of the images. 
This means that for practical inference on larger areas, the rasters should be split into patches with an overlap of 1/4 of the image size between images in each dimension. 
Only the model using the sole RE input showed outlying performance, which could again be related to the model overfitting less to the undesired features, as was presumed to be the case for the coarse pseudo-label trained models.

\subsection{Recommendations and future directions}
\label{sec:future_work}
To further leverage our method to produce tree crown instance segmentation models that would perform better in any given scenario, we believe the data should be combined with similar datasets from more varied geographic areas and time periods. 
Additionally, the model could be improved, for example, by changing from a Mask R-CNN to something which has been shown to work better on tree crowns specifically, such as the StarDist model~\citep{tong2025individual}, 
by simply using a more modern backbone architecture, such as a Vision Transformer~\citep{dosovitskiy2020image}, or by focusing on fine-tuning a model with improved pretraining akin to DINOv3~\citep{simeoni2025dinov3} or SAM 3~\citep{carion2025sam3segmentconcepts}.

For future development, it would also likely be possible to obtain similar, domain-specific results using a far more efficient model, such as a small version of PIDNet~\citep{xu2023pidnet}. In contrast, for work on more generalist tree crown segmentation models, it might be beneficial to lean towards models with parameter counts similar to that of SAM 3.
Finally, during the first experiments, the more recent SAM 3 was not yet available and, as such, it was not used for the pseudo-label generation process, but we presume it could yield similar results with most likely slight improvements.

\section{Conclusion}
\label{sec:conclusion}

Our work shows that training domain-specific models for segmenting individual tree crowns from aerial images is feasible using automatically generated CHM-based pseudo-labels without manual training annotations. 
Importantly, the performance of our proposed models fell within the range reported for fully supervised models in previous studies under comparable conditions. 
Finally, incorporating information from SAM 2 to enhance the pseudo-labels substantially improved downstream segmentation performance, increasing the best-performing model mIoU by 9.9\% and the average mIoU across all models by 8.4\%. These results demonstrate the potential of combining automatically generated remote-sensing labels with foundation models to efficiently train domain-specific tree crown segmentation models. Furthermore, the study demonstrates the potential of reusing older ALS data to generate training annotations for more recent aerial imagery.

\section*{Acknowledgements}

The authors wish to acknowledge CSC – IT Center for Science, Finland, for computational resources. 
We thank Ville Luoma, Tuomas Yrttimaa, Jiri Pyörälä, Osmo Suominen, and Otto Saikkonen for collecting the manual measurements of the Evo data used to fit the Näslund function.
We also thank Valtteri Soininen for the idea of using the Näslund function to estimate the tree heights from the diameter information. We thank Svenska Cellulosa Aktiebolaget (SCA) for providing access to the Sundsvall test site and the associated harvester reference data.

\section*{Declaration of competing interest}

The authors declare no competing interests.

\section*{Data availability}

Data release is planned for a later date due to ongoing research.

\section*{Funding sources}
This research was funded by the Research Council of Finland (RCF) within projects “Learning techniques for autonomous drone-based hyperspectral analysis of forest vegetation” (decision no. 357380) and RCF research Infrastructure “Measuring Spatiotemporal Changes in Forest Ecosystem” (346382), as well as the European Union – NextGenerationEU instrument through the Research Council of Finland (grant number 353263 for Multirisk Project). Additional support was provided by the SingleTree project, funded by the Circular Bio-based Europe Joint Undertaking (CBE JU) and its members under Grant Agreement No. 101157488. This study has been performed with affiliation to the RCF Flagship Forest–Human–Machine Interplay—Building Resilience, Redefining Value Networks and Enabling Meaningful Experiences (UNITE) (decision no. 357908).

Funded by the European Union. Views and opinions expressed are, however, those of the author(s) only and do not necessarily reflect those of the European Union or CBE JU. Neither the European Union nor the CBE JU can be held responsible for them.

\section*{CRediT authorship contribution statement}

\textbf{Julius Pesonen:} Conceptualisation, Methodology, Software, Writing - Original Draft, Validation, Visualisation.
\textbf{Stefan Rua:} Software, Methodology, Data Curation, Writing - Original Draft, Visualisation.
\textbf{Josef Taher:} Resources, Writing - Review \& Editing.
\textbf{Niko Koivumäki:} Resources, Writing - Review \& Editing.
\textbf{Xiaowei Yu:} Software, Resources.
\textbf{Teemu Hakala:} Resources.
\textbf{Eija Honkavaara:} Supervision, Funding acquisition, Writing - Review \& Editing.
\bibliographystyle{elsarticle-harv}
\bibliography{bib}

@article{chen2025zero,
  title={Zero-Shot Tree Detection and Segmentation from Aerial Forest Imagery},
  author={Chen, Michelle and Russell, David and Pallavoor, Amritha and Young, Derek and Wu, Jane},
  journal={arXiv:2506.03114},
  year={2025},
doi={https://doi.org/10.48550/arXiv.2506.03114},
note={[preprint]},
}

@article{zhu2025leveraging,
  title={Leveraging {SAM 2} and {LiDAR} for Automated Individual Tree Crown Delineation: A Comparative Evaluation of Prompting Methods},
  author={Zhu, Yun and Locke, William and Yuan, Jingyi and Zhang, Yunqian and Ma, Qin and Liang, Lu},
  journal={Information Geography},
  pages={100025},
  year={2025},
  publisher={Elsevier},
doi={https://doi.org/10.1016/j.infgeo.2025.100025},
}

@article{weinstein2020deepforest,
  title={{DeepForest}: A Python package for {RGB} deep learning tree crown delineation},
  author={Weinstein, Ben G and Marconi, Sergio and Aubry-Kientz, M{\'e}laine and Vincent, Gregoire and Senyondo, Henry and White, Ethan P},
  journal={Methods in Ecology and Evolution},
  volume={11},
  number={12},
  pages={1743--1751},
  year={2020},
  publisher={Wiley Online Library},
doi={https://doi.org/10.1111/2041-210X.13472},
}

@article{ball2023accurate,
  title={Accurate delineation of individual tree crowns in tropical forests from aerial {RGB} imagery using {Mask R-CNN}},
  author={Ball, James GC and Hickman, Sebastian HM and Jackson, Tobias D and Koay, Xian Jing and Hirst, James and Jay, William and Archer, Matthew and Aubry-Kientz, M{\'e}laine and Vincent, Gr{\'e}goire and Coomes, David A},
  journal={Remote Sensing in Ecology and Conservation},
  volume={9},
  number={5},
  pages={641--655},
  year={2023},
  publisher={Wiley Online Library},
doi={https://doi.org/10.1002/rse2.332},
}

@article{kirillov2023segany,
  title={Segment Anything}, 
  author={Kirillov, Alexander and Mintun, Eric and Ravi, Nikhila and Mao, Hanzi and Rolland, Chloe and Gustafson, Laura and Xiao, Tete and Whitehead, Spencer and Berg, Alexander C. and Lo, Wan-Yen and Doll{\'a}r, Piotr and Girshick, Ross},
  journal={arXiv:2304.02643},
  year={2023},
doi={https://doi.org/10.48550/arXiv.2304.02643},
note={[preprint]},
}

@article{liu2023grounding,
  title={Grounding {DINO}: Marrying {DINO} with Grounded Pre-Training for Open-Set Object Detection},
  author={Liu, Shilong and Zeng, Zhaoyang and Ren, Tianhe and Li, Feng and Zhang, Hao and Yang, Jie and Li, Chunyuan and Yang, Jianwei and Su, Hang and Zhu, Jun and others},
  journal={arXiv:2303.05499},
  year={2023},
    doi={https://doi.org/10.48550/arXiv.2303.05499},
note={[preprint]}
}

@misc{ren2024grounded,
      title={{Grounded SAM}: Assembling Open-World Models for Diverse Visual Tasks}, 
      author={Tianhe Ren and Shilong Liu and Ailing Zeng and Jing Lin and Kunchang Li and He Cao and Jiayu Chen and Xinyu Huang and Yukang Chen and Feng Yan and Zhaoyang Zeng and Hao Zhang and Feng Li and Jie Yang and Hongyang Li and Qing Jiang and Lei Zhang},
      year={2024},
      eprint={2401.14159},
      archivePrefix={arXiv},
      primaryClass={cs.CV},
    doi={https://doi.org/10.48550/arXiv.2401.14159},
note={[preprint]}
}

@inproceedings{arnaudo2024fmars,
  title={{FMARS}: Annotating remote sensing images for disaster management using foundation models},
  author={Arnaudo, Edoardo and Vaschetti, Jacopo Lungo and Innocenti, Lorenzo and Barco, Luca and Lisi, Davide and Fissore, Vanina and Rossi, Claudio},
  booktitle={IGARSS 2024-2024 IEEE International Geoscience and Remote Sensing Symposium},
  pages={3920--3924},
  year={2024},
  organization={IEEE},
doi={https://doi.org/10.1109/IGARSS53475.2024.10641130},
}

@article{teng2025assessing,
  title={Assessing {SAM} for Tree Crown Instance Segmentation from Drone Imagery},
  author={Teng, M{\'e}lisande and Ouaknine, Arthur and Lalibert{\'e}, Etienne and Bengio, Yoshua and Rolnick, David and Larochelle, Hugo},
  journal={arXiv:2503.20199},
  year={2025},
doi={https://doi.org/10.48550/arXiv.2503.20199},
note={[preprint]}
}

@article{chen2024rsprompter,
  title={{RSPrompter}: Learning to prompt for remote sensing instance segmentation based on visual foundation model},
  author={Chen, Keyan and Liu, Chenyang and Chen, Hao and Zhang, Haotian and Li, Wenyuan and Zou, Zhengxia and Shi, Zhenwei},
  journal={IEEE Transactions on Geoscience and Remote Sensing},
  volume={62},
  pages={1--17},
  year={2024},
  publisher={IEEE},
doi={https://doi.org/10.1109/TGRS.2024.3356074},
}

@inproceedings{khan2025validation,
  title={Validation Challenges in Large-Scale Tree Crown Segmentations from Remote Sensing Imagery Using Deep Learning: A Case Study in Germany},
  author={Khan, Taimur and Krebs, Jasmin and Gupta, Sharad Kumar and Renkel, Jonathan and Arnold, Caroline and N{\"o}lke, Nils},
  booktitle={International Conference on Theory and Practice of Digital Libraries},
  pages={311--323},
  year={2025},
  organization={Springer},
doi={https://doi.org/10.1007/978-3-032-06136-2\_30},
}

@article{tong2025individual,
  title={Individual tree crown delineation in high resolution aerial {RGB} imagery using {StarDist}-based model},
  author={Tong, Fei and Zhang, Yun},
  journal={Remote Sensing of Environment},
  volume={319},
  pages={114618},
  year={2025},
  publisher={Elsevier},
doi={https://doi.org/10.1016/j.rse.2025.114618},
}

@article{dersch2024semi,
  title={Semi-supervised multi-class tree crown delineation using aerial multispectral imagery and lidar data},
  author={Dersch, S and Sch{\"o}ttl, A and Krzystek, P and Heurich, M},
  journal={ISPRS Journal of Photogrammetry and Remote Sensing},
  volume={216},
  pages={154--167},
  year={2024},
  publisher={Elsevier},
doi={https://doi.org/10.1016/j.isprsjprs.2024.07.032},
}

@article{zheng2024review,
  title={A review of individual tree crown detection and delineation from optical remote sensing images: Current progress and future},
  author={Zheng, Juepeng and Yuan, Shuai and Li, Weijia and Fu, Haohuan and Yu, Le and Huang, Jianxi},
  journal={IEEE Geoscience and Remote Sensing Magazine},
  year={2024},
  publisher={IEEE},
doi={https://doi.org/10.1109/MGRS.2024.3479871},
}

@article{freudenberg2022individual,
  title={Individual tree crown delineation in high-resolution remote sensing images based on {U-Net}},
  author={Freudenberg, Maximilian and Magdon, Paul and Nölke, Nils},
  journal={NCAA},
  year={2022},
  publisher={Springer},
  doi={https://doi.org/10.1007/s00521-022-07640-4}
}

@inproceedings{
ravi2024sam2segmentimages,
title={{SAM} 2: Segment Anything in Images and Videos},
author={Nikhila Ravi and Valentin Gabeur and Yuan-Ting Hu and Ronghang Hu and Chaitanya Ryali and Tengyu Ma and Haitham Khedr and Roman R{\"a}dle and Chloe Rolland and Laura Gustafson and Eric Mintun and Junting Pan and Kalyan Vasudev Alwala and Nicolas Carion and Chao-Yuan Wu and Ross Girshick and Piotr Dollar and Christoph Feichtenhofer},
booktitle={The Thirteenth International Conference on Learning Representations},
year={2025},
url={https://openreview.net/forum?id=Ha6RTeWMd0}
}

@misc{carion2025sam3segmentconcepts,
      title={{SAM} 3: Segment Anything with Concepts},
      author={Nicolas Carion and Laura Gustafson and Yuan-Ting Hu and Shoubhik Debnath and Ronghang Hu and Didac Suris and Chaitanya Ryali and Kalyan Vasudev Alwala and Haitham Khedr and Andrew Huang and Jie Lei and Tengyu Ma and Baishan Guo and Arpit Kalla and Markus Marks and Joseph Greer and Meng Wang and Peize Sun and Roman Rädle and Triantafyllos Afouras and Effrosyni Mavroudi and Katherine Xu and Tsung-Han Wu and Yu Zhou and Liliane Momeni and Rishi Hazra and Shuangrui Ding and Sagar Vaze and Francois Porcher and Feng Li and Siyuan Li and Aishwarya Kamath and Ho Kei Cheng and Piotr Dollár and Nikhila Ravi and Kate Saenko and Pengchuan Zhang and Christoph Feichtenhofer},
      year={2025},
      eprint={2511.16719},
      archivePrefix={arXiv},
      primaryClass={cs.CV},
doi={https://doi.org/10.48550/arXiv.2511.16719},
note={[preprint]},
}

@article{lin2024individual,
  title={Individual Tree Crown Delineation Using Airborne LiDAR Data and Aerial Imagery in the Taiga--Tundra Ecotone},
  author={Lin, Yuanyuan and Li, Hui and Jing, Linhai and Ding, Haifeng and Tian, Shufang},
  journal={Remote Sensing},
  volume={16},
  number={21},
  pages={3920},
  year={2024},
  publisher={MDPI},
doi={https://doi.org/10.3390/rs16213920},
}

@misc{simeoni2025dinov3,
  title={{DINOv3}},
  author={Sim{\'e}oni, Oriane and Vo, Huy V. and Seitzer, Maximilian and Baldassarre, Federico and Oquab, Maxime and Jose, Cijo and Khalidov, Vasil and Szafraniec, Marc and Yi, Seungeun and Ramamonjisoa, Micha{\"e}l and Massa, Francisco and Haziza, Daniel and Wehrstedt, Luca and Wang, Jianyuan and Darcet, Timoth{\'e}e and Moutakanni, Th{\'e}o and Sentana, Leonel and Roberts, Claire and Vedaldi, Andrea and Tolan, Jamie and Brandt, John and Couprie, Camille and Mairal, Julien and J{\'e}gou, Herv{\'e} and Labatut, Patrick and Bojanowski, Piotr},
  year={2025},
  journal={arXiv:2508.10104},
  primaryClass={cs.CV},
  url={https://arxiv.org/abs/2508.10104},
  note={[preprint]}
}

@inproceedings{
dosovitskiy2020image,
title={An Image is Worth 16x16 Words: Transformers for Image Recognition at Scale},
author={Alexey Dosovitskiy and Lucas Beyer and Alexander Kolesnikov and Dirk Weissenborn and Xiaohua Zhai and Thomas Unterthiner and Mostafa Dehghani and Matthias Minderer and Georg Heigold and Sylvain Gelly and Jakob Uszkoreit and Neil Houlsby},
booktitle={International Conference on Learning Representations},
year={2021},
url={https://openreview.net/forum?id=YicbFdNTTy}
}

@article{karila2019effect,
  title={The effect of seasonal variation on automated land cover mapping from multispectral airborne laser scanning data},
  author={Karila, Kirsi and Matikainen, Leena and Litkey, Paula and Hyypp{\"a}, Juha and Puttonen, Eetu},
  journal={International Journal of Remote Sensing},
  volume={40},
  number={9},
  pages={3289--3307},
  year={2019},
  publisher={Taylor \& Francis},
doi={https://doi.org/10.1080/01431161.2018.1528023},
}

@inproceedings{he2017mask,
  title={{Mask R-CNN}},
  author={He, Kaiming and Gkioxari, Georgia and Doll{\'a}r, Piotr and Girshick, Ross},
  booktitle={Proceedings of the IEEE international conference on computer vision},
  pages={2961--2969},
  year={2017},
doi={https://doi.org/10.1109/ICCV.2017.322},
}

@inproceedings{deng2009imagenet,
  title={{ImageNet}: A large-scale hierarchical image database},
  author={Deng, Jia and Dong, Wei and Socher, Richard and Li, Li-Jia and Li, Kai and Fei-Fei, Li},
  booktitle={2009 IEEE conference on computer vision and pattern recognition},
  pages={248--255},
  year={2009},
  organization={Ieee},
doi={https://doi.org/10.1109/CVPR.2009.5206848},
}

@inproceedings{he2016deep,
  title={Deep residual learning for image recognition},
  author={He, Kaiming and Zhang, Xiangyu and Ren, Shaoqing and Sun, Jian},
  booktitle={Proceedings of the IEEE conference on computer vision and pattern recognition},
  pages={770--778},
  year={2016},
doi={https://doi.ieeecomputersociety.org/10.1109/CVPR.2016.90},
}

@article{li2021benchmarking,
  title={Benchmarking detection transfer learning with vision transformers},
  author={Li, Yanghao and Xie, Saining and Chen, Xinlei and Dollar, Piotr and He, Kaiming and Girshick, Ross},
  journal={arXiv:2111.11429},
  year={2021},
doi={https://doi.org/10.48550/arXiv.2111.11429},
note={[preprint]},
}

@inproceedings{lin2017feature,
  title={Feature pyramid networks for object detection},
  author={Lin, Tsung-Yi and Doll{\'a}r, Piotr and Girshick, Ross and He, Kaiming and Hariharan, Bharath and Belongie, Serge},
  booktitle={Proceedings of the IEEE conference on computer vision and pattern recognition},
  pages={2117--2125},
  year={2017},
doi={https://doi.org/10.1109/CVPR.2017.106},
}

@inproceedings{
loshchilov2017decoupled,
title={Decoupled Weight Decay Regularization},
author={Ilya Loshchilov and Frank Hutter},
booktitle={International Conference on Learning Representations},
year={2019},
url={https://openreview.net/forum?id=Bkg6RiCqY7},
}

@article{kingma2014adam,
  title={Adam: A method for stochastic optimization},
  author={Kingma, Diederik P},
  journal={arXiv:1412.6980},
  year={2014},
doi={https://doi.org/10.48550/arXiv.1412.6980},
note={[preprint]},
}

@inproceedings{hanson1988comparing,
 author = {Hanson, Stephen and Pratt, Lorien},
 booktitle = {Advances in Neural Information Processing Systems},
 editor = {D. Touretzky},
 pages = {},
 publisher = {Morgan-Kaufmann},
 title = {Comparing Biases for Minimal Network Construction with Back-Propagation},
 volume = {1},
 year = {1988}
}

@inproceedings{liu2016ssd,
  title={{SSD}: Single shot multibox detector},
  author={Liu, Wei and Anguelov, Dragomir and Erhan, Dumitru and Szegedy, Christian and Reed, Scott and Fu, Cheng-Yang and Berg, Alexander C},
  booktitle={European conference on computer vision},
  pages={21--37},
  year={2016},
  organization={Springer},
doi={https://doi.org/10.1007/978-3-319-46448-0\_2},
}

@misc{ball_2025_15863800,
  author       = {Ball, James and
                  Kotthoff, Christopher},
  title        = {Detectree2 trained models},
  year         = 2025,
  publisher    = {Zenodo},
  doi          = {10.5281/zenodo.15863800},
  url          = {https://doi.org/10.5281/zenodo.15863800},
}

@article {Ball2024.06.24.600405,
	author = {Ball, James G C and Jaffer, Sadiq and Laybros, Anthony and Prieur, Colin and Jackson, Toby and Madhavapeddy, Anil and Barbier, Nicolas and Vincent, Gregoire and Coomes, David A},
	title = {Towards comprehensive individual tree species mapping in diverse tropical forests by harnessing temporal and spectral dimensions},
	elocation-id = {2024.06.24.600405},
	year = {2025},
	doi = {https://doi.org/10.1101/2024.06.24.600405 },
	publisher = {Cold Spring Harbor Laboratory},
	journal = {bioRxiv},
note={[preprint]},
}

@article{lai2019impact,
  title={The impact of urban street tree species on air quality and respiratory illness: A spatial analysis of large-scale, high-resolution urban data},
  author={Lai, Yuan and Kontokosta, Constantine E},
  journal={Health \& place},
  volume={56},
  pages={80--87},
  year={2019},
  publisher={Elsevier},
doi = {https://doi.org/10.1016/j.healthplace.2019.01.016},
}

@article{you2017geographical,
  title={Geographical information system-based forest fire risk assessment integrating national forest inventory data and analysis of its spatiotemporal variability},
  author={You, Weibin and Lin, Li and Wu, Liyun and Ji, Zhirong and Yu, Jian’an and Zhu, Jianqin and Fan, Yunjian and He, Dongjin},
  journal={Ecological Indicators},
  volume={77},
  pages={176--184},
  year={2017},
  publisher={Elsevier},
doi={https://doi.org/10.1016/j.ecolind.2017.01.042},
}

@article{zhao2015quantifying,
  title={Quantifying and mapping the supply of and demand for carbon storage and sequestration service from urban trees},
  author={Zhao, Chang and Sander, Heather A},
  journal={PLoS One},
  volume={10},
  number={8},
  pages={e0136392},
  year={2015},
  publisher={Public Library of Science San Francisco, CA USA},
doi={https://doi.org/10.1371/journal.pone.0136392},
}

@article{10.5849/forsci.12-134,
    author = {Brosofske, Kimberley D. and Froese, Robert E. and Falkowski, Michael J. and Banskota, Asim},
    title = {A Review of Methods for Mapping and Prediction of Inventory Attributes for Operational Forest Management},
    journal = {Forest Science},
    volume = {60},
    number = {4},
    pages = {733-756},
    year = {2013},
    month = {11},
    issn = {0015-749X},
    doi = {{https://doi.org/10.5849/forsci.12-134}},
}

@article{patacca2023significant,
  title={Significant increase in natural disturbance impacts on European forests since 1950},
  author={Patacca, Marco and Lindner, Marcus and Lucas-Borja, Manuel Esteban and Cordonnier, Thomas and Fidej, Gal and Gardiner, Barry and Hauf, Ylva and Jasinevi{\v{c}}ius, Gediminas and Labonne, Sophie and Linkevi{\v{c}}ius, Edgaras and others},
  journal={Global change biology},
  volume={29},
  number={5},
  pages={1359--1376},
  year={2023},
  publisher={Wiley Online Library},
doi={https://doi.org/10.1111/gcb.16531},
}

@article{hlasny2021bark,
  title={Bark beetle outbreaks in Europe: state of knowledge and ways forward for management},
  author={Hl{\'a}sny, Tom{\'a}{\v{s}} and K{\"o}nig, Louis and Krokene, Paal and Lindner, Marcus and Montagn{\'e}-Huck, Claire and M{\"u}ller, J{\"o}rg and Qin, Hua and Raffa, Kenneth F and Schelhaas, Mart-Jan and Svoboda, Miroslav and others},
  journal={Current Forestry Reports},
  volume={7},
  number={3},
  pages={138--165},
  year={2021},
  publisher={Springer},
doi={https://doi.org/10.1007/s40725-021-00142-x},
}

@article{yang2025mapping,
  title={Mapping carbon stock and growth of individual street trees using LiDAR-camera fusion-based mobile mapping system},
  author={Yang, Tackang and Ryu, Youngryel and Kwon, Ryoungseob and Choi, Changhyun and Zhong, Zilong and Nam, Yunsoo and Jo, Seongwoo},
  journal={Remote Sensing of Environment},
  volume={328},
  pages={114895},
  year={2025},
  publisher={Elsevier},
doi={https://doi.org/10.1016/j.rse.2025.114895},
}

@misc{usda_urban_inv,
    title={Urban Forest Inventory and Analysis Program},
    author={{Forest Service, U.S. Department of Agriculture}},
    available={https://research.fs.usda.gov/programs/urbanfia},
    year={2025},
    note={Available at: https://research.fs.usda.gov/programs/urbanfia, (accessed 12 January 2026)},
}

@article{zhou2013mapping,
  title={Mapping local density of young Eucalyptus plantations by individual tree detection in high spatial resolution satellite images},
  author={Zhou, Jia and Proisy, Christophe and Descombes, Xavier and Le Maire, Guerric and Nouvellon, Yann and Stape, Jose-Luiz and Viennois, Ga{\"e}lle and Zerubia, Josiane and Couteron, Pierre},
  journal={Forest Ecology and Management},
  volume={301},
  pages={129--141},
  year={2013},
  publisher={Elsevier},
doi={https://doi.org/10.1016/j.foreco.2012.10.007},
}

@article{zhang2023mapping,
  title={A mapping approach for eucalyptus plantations canopy and single tree using high-resolution satellite images in {Liuzhou, China}},
  author={Zhang, Sen and Cui, Yaoping and Zhou, Yan and Dong, Junwu and Li, Wanlong and Liu, Bailu and Dong, Jinwei},
  journal={IEEE Transactions on Geoscience and Remote Sensing},
  volume={61},
  pages={1--13},
  year={2023},
  publisher={IEEE},
doi={https://doi.org/10.1109/TGRS.2023.3327128},
}

@article{wolk2024review,
  title={A review of semantic segmentation and instance segmentation techniques in forestry using LiDAR and imagery data},
  author={Wo{\l}k, Krzysztof and Tatara, Marek S},
  journal={Electronics},
  volume={13},
  number={20},
  pages={4139},
  year={2024},
  publisher={MDPI},
doi={https://doi.org/10.3390/electronics13204139},
}

@article{aubry2021multisensor,
  title={Multisensor data fusion for improved segmentation of individual tree crowns in dense tropical forests},
  author={Aubry-Kientz, M{\'e}laine and Laybros, Anthony and Weinstein, Ben and Ball, James GC and Jackson, Toby and Coomes, David and Vincent, Gr{\'e}goire},
  journal={IEEE Journal of Selected Topics in Applied Earth Observations and Remote Sensing},
  volume={14},
  pages={3927--3936},
  year={2021},
  publisher={IEEE},
doi={https://doi.org/10.1109/JSTARS.2021.3069159},
}

@incollection{favorskaya2015fusion,
  title={Fusion of airborne LiDAR and digital photography data for tree crowns segmentation and measurement},
  author={Favorskaya, Margarita and Tkacheva, Anastasia and Danilin, Igor M and Medvedev, Evgeny M},
  booktitle={Intelligent interactive multimedia systems and services},
  pages={191--201},
  year={2015},
  publisher={Springer},
doi={https://doi.org/10.1007/978-3-319-19830-9\_18},
}

@mastersthesis{rua2025sam2,
  title={{SAM}2 Pseudolabeling for Instance Segmentation of Tree Crowns from Aerial Imagery},
  author={Rua, Stefan},
  year={2025},
school={Aalto University},
url={https://urn.fi/URN:NBN:fi:aalto-202512179400},
}

@article{taher2026multispectral,
  title={Multispectral airborne laser scanning for tree species classification: a benchmark of machine learning and deep learning algorithms},
  author={Taher, Josef and Hyypp{\"a}, Eric and Hyypp{\"a}, Matti and Salolahti, Klaara and Yu, Xiaowei and Matikainen, Leena and Kukko, Antero and Lehtom{\"a}ki, Matti and Kaartinen, Harri and Thurachen, Sopitta and others},
  journal={ISPRS Journal of Photogrammetry and Remote Sensing},
  volume={233},
  pages={278--309},
  year={2026},
  publisher={Elsevier},
doi={https://doi.org/10.1016/j.isprsjprs.2026.01.031},
}

@article{kaartinen2012international,
  title={An international comparison of individual tree detection and extraction using airborne laser scanning},
  author={Kaartinen, Harri and Hyypp{\"a}, Juha and Yu, Xiaowei and Vastaranta, Mikko and Hyypp{\"a}, Hannu and Kukko, Antero and Holopainen, Markus and Heipke, Christian and Hirschmugl, Manuela and Morsdorf, Felix and others},
  journal={Remote Sensing},
  volume={4},
  number={4},
  pages={950--974},
  year={2012},
  publisher={Molecular Diversity Preservation International},
doi={https://doi.org/10.3390/rs4040950},
}

@article{matikainen2017object,
  title={Object-based analysis of multispectral airborne laser scanner data for land cover classification and map updating},
  author={Matikainen, Leena and Karila, Kirsi and Hyypp{\"a}, Juha and Litkey, Paula and Puttonen, Eetu and Ahokas, Eero},
  journal={ISPRS Journal of Photogrammetry and Remote Sensing},
  volume={128},
  pages={298--313},
  year={2017},
  publisher={Elsevier},
doi={https://doi.org/10.1016/j.isprsjprs.2017.04.005},
}

@misc{agisoft,
    author = {Agisoft},
    title = {{Agisoft Metashape Professional}},
    note = {(v.2.0.2). [software], https://www.agisoft.com/},
    year = {2023},
}

@misc{lastools,
    author = {rapidlasso GmbH},
    title = {{LAStools} software suite},
    note = {[software], https://rapidlasso.de/},
    year = {2024},
}

@article{brandt2026chmv2,
  title={{CHMv2}: Improvements in Global Canopy Height Mapping using {DINOv3}},
  author={Brandt, John and Yi, Seungeun and Tolan, Jamie and Li, Xinyuan and Potapov, Peter and Ertel, Jessica and Spore, Justine and Vo, Huy V and Ramamonjisoa, Micha{\"e}l and Labatut, Patrick and others},
  doi={https://doi.org/10.48550/arXiv.2603.06382},
  journal={arXiv:2603.06382},
  year={2026},
  note={[preprint]}
}

@article{tolan2024very,
  title={Very high resolution canopy height maps from {RGB} imagery using self-supervised vision transformer and convolutional decoder trained on aerial lidar},
  author={Tolan, Jamie and Yang, Hung-I and Nosarzewski, Benjamin and Couairon, Guillaume and Vo, Huy V and Brandt, John and Spore, Justine and Majumdar, Sayantan and Haziza, Daniel and Vamaraju, Janaki and others},
  journal={Remote Sensing of Environment},
  volume={300},
  pages={113888},
  year={2024},
  publisher={Elsevier},
  doi={https://doi.org/10.1016/j.rse.2023.113888}
}

@inproceedings{xu2023pidnet,
  title={{PIDNet}: A real-time semantic segmentation network inspired by {PID} controllers},
  author={Xu, Jiacong and Xiong, Zixiang and Bhattacharyya, Shankar P},
  booktitle={Proceedings of the IEEE/CVF conference on computer vision and pattern recognition},
  pages={19529--19539},
  year={2023},
  doi={https://doi.org/10.1109/CVPR52729.2023.01871}
}

@article{blanco2020potential,
  title={Potential of {UAS}-based remote sensing for estimating tree water status and yield in sweet cherry trees},
  author={Blanco, V{\'\i}ctor and Blaya-Ros, Pedro Jos{\'e} and Castillo, Cristina and Soto-Vall{\'e}s, Fulgencio and Torres-S{\'a}nchez, Roque and Domingo, Rafael},
  journal={Remote Sensing},
  volume={12},
  number={15},
  pages={2359},
  year={2020},
  doi={https://doi.org/10.3390/rs12152359},
  publisher={MDPI}
}

@article{hashim2019urban,
  title={Urban vegetation classification with {NDVI} threshold value method with very high resolution ({VHR}) Pleiades imagery},
  author={Hashim, Haslina and Abd Latif, Zulkiflee and Adnan, Nor Aizam},
  journal={The International Archives of the Photogrammetry, Remote Sensing and Spatial Information Sciences},
  volume={42},
  pages={237--240},
  year={2019},
  doi={https://doi.org/10.5194/isprs-archives-XLII-4-W16-237-2019},
  publisher={Copernicus GmbH}
}

@article{bhandari2012feature,
  title={Feature extraction using Normalized Difference Vegetation Index ({NDVI}): A case study of Jabalpur city},
  author={Bhandari, AK and Kumar, A and Singh, GK},
  journal={Procedia technology},
  volume={6},
  pages={612--621},
  year={2012},
  doi = {https://doi.org/10.1016/j.protcy.2012.10.074},
  publisher={Elsevier}
}

@dataset{locke_2025_15420565,
  author       = {Locke, William},
  title        = {Subset of {NeonTreeEvaluation} Dataset with Manual
                   Deliniations
                  },
  month        = May,
  year         = 2025,
  publisher    = {Zenodo},
  doi          = {10.5281/zenodo.15420565},
  url          = {https://doi.org/10.5281/zenodo.15420565},
}

@article{troles2024bamforests,
  title={{BAMFORESTS}: Bamberg benchmark forest dataset of individual tree crowns in very-high-resolution {UAV} images},
  author={Troles, Jonas and Schmid, Ute and Fan, Wen and Tian, Jiaojiao},
  journal={Remote Sensing},
  volume={16},
  number={11},
  pages={1935},
  year={2024},
  publisher={MDPI},
  doi={https://doi.org/10.3390/rs16111935}
}

@article{okhrimenko2019multi,
  title={Multi-spectral lidar: Radiometric calibration, canopy spectral reflectance, and vegetation vertical {SVI} profiles},
  author={Okhrimenko, Maxim and Coburn, Craig and Hopkinson, Chris},
  journal={Remote Sensing},
  volume={11},
  number={13},
  pages={1556},
  year={2019},
  publisher={MDPI},
  doi={https://doi.org/10.3390/rs11131556}
}

@article{minaee2021image,
  title={Image segmentation using deep learning: A survey},
  author={Minaee, Shervin and Boykov, Yuri and Porikli, Fatih and Plaza, Antonio and Kehtarnavaz, Nasser and Terzopoulos, Demetri},
  journal={IEEE transactions on pattern analysis and machine intelligence},
  volume={44},
  number={7},
  pages={3523--3542},
  year={2021},
  doi={https://doi.org/10.1109/TPAMI.2021.3059968},
  publisher={IEEE}
}

@article{zhen2016trends,
  title={Trends in automatic individual tree crown detection and delineation—Evolution of LiDAR data},
  author={Zhen, Zhen and Quackenbush, Lindi J and Zhang, Lianjun},
  journal={Remote Sensing},
  volume={8},
  number={4},
  pages={333},
  year={2016},
  publisher={MDPI},
  doi={https://doi.org/10.3390/rs8040333}
}

@article{hakula2023individual,
  title={Individual tree segmentation and species classification using high-density close-range multispectral laser scanning data},
  author={Hakula, Aada and Ruoppa, Lassi and Lehtom{\"a}ki, Matti and Yu, Xiaowei and Kukko, Antero and Kaartinen, Harri and Taher, Josef and Matikainen, Leena and Hyypp{\"a}, Eric and Luoma, Ville and others},
  journal={ISPRS Open Journal of Photogrammetry and Remote Sensing},
  volume={9},
  pages={100039},
  year={2023},
  publisher={Elsevier}
}

@book{b603dea5c4a74aa08d50ffedad328bbc,
title = "Skogsf{\"o}rs{\"o}ksanstaltens gallringsf{\"o}rs{\"o}k i tallskog: prim{\"a}rbearbetning",
author = "Manfred N{\"a}slund",
year = "1936",
language = "Svenska",
series = "Meddelanden fr{\aa}n Statens skogsf{\"o}rs{\"o}ksanstalt",
number = "29:1",
publisher = "Statens skogsf{\"o}rs{\"o}ksanstalt",

}

@article{HYYPPA2026100142,
title = {Multispectral airborne laser scanning dataset for tree species classification: MS-ALS-SPECIES},
journal = {ISPRS Open Journal of Photogrammetry and Remote Sensing},
volume = {21},
pages = {100142},
year = {2026},
issn = {2667-3932},
doi = {https://doi.org/10.1016/j.ophoto.2026.100142},
url = {https://www.sciencedirect.com/science/article/pii/S2667393226000281},
author = {Matti Hyyppä and Klaara Salolahti and Eric Hyyppä and Xiaowei Yu and Josef Taher and Leena Matikainen and Matti Lehtomäki and Paula Litkey and Teemu Hakala and Harri Kaartinen and Juha Hyyppä and Antero Kukko},
}

@article{weinstein2019individual,
  title={Individual tree-crown detection in {RGB} imagery using semi-supervised deep learning neural networks},
  author={Weinstein, Ben G and Marconi, Sergio and Bohlman, Stephanie and Zare, Alina and White, Ethan},
  journal={Remote Sensing},
  volume={11},
  number={11},
  pages={1309},
  year={2019},
  publisher={MDPI}
}

@misc{https://doi.org/10.48443/xv9v-1r95,
  doi = {10.48443/XV9V-1R95},
  url = {https://data.neonscience.org/data-products/DP1.30010.001/RELEASE-2026},
  author = {{National Ecological Observatory Network (NEON)}},
  language = {en},
  title = {High-resolution orthorectified camera imagery (DP1.30010.001)},
  publisher = {National Ecological Observatory Network (NEON)},
  year = {2026}
}

@article{ZHENG2022103074,
title = {Individual tree-based vs pixel-based approaches to mapping forest functional traits and diversity by remote sensing},
journal = {International Journal of Applied Earth Observation and Geoinformation},
volume = {114},
pages = {103074},
year = {2022},
issn = {1569-8432},
doi = {https://doi.org/10.1016/j.jag.2022.103074},
url = {https://www.sciencedirect.com/science/article/pii/S156984322200262X},
author = {Zhaoju Zheng and Yuan Zeng and Meredith C. Schuman and Hailan Jiang and Bernhard Schmid and Michael E. Schaepman and Felix Morsdorf},
}

\clearpage
\appendix
\section{Additional Figures}
\label{app:figures}

\setcounter{figure}{0}

\begin{figure}[!ht]
    \centering
    \includegraphics[width=\linewidth]{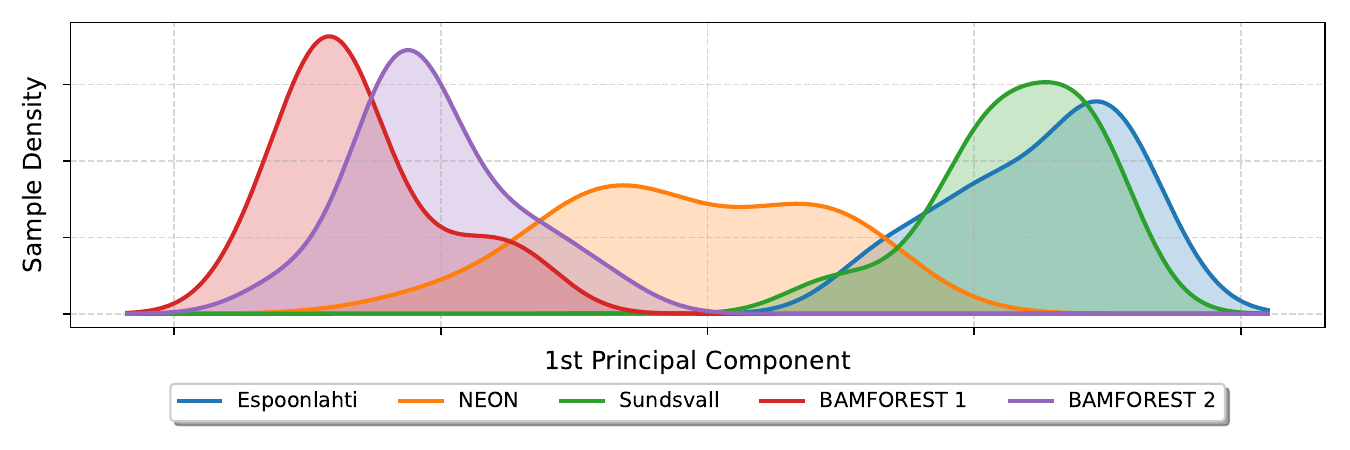}
    \caption{Distributions of the different datasets used in the study, highlighting the large shift between public datasets (BAMFOREST and NEON) and data from our targeted deployment regions (Espoonlahti and Sundsvall). 
    Distributions are visualised based on sample density over the 1st principal component computed from individual image features extracted using an ImageNet~\citep{deng2009imagenet} pretrained ViT model~\citep{dosovitskiy2020image}.
    The figure is illustrative, but it roughly corresponds to the qualitative visual differences between the datasets.}
    \label{fig:dataset_features}
\end{figure}

\begin{figure}[!ht]
    \centering
    \includegraphics[width=0.8\linewidth]{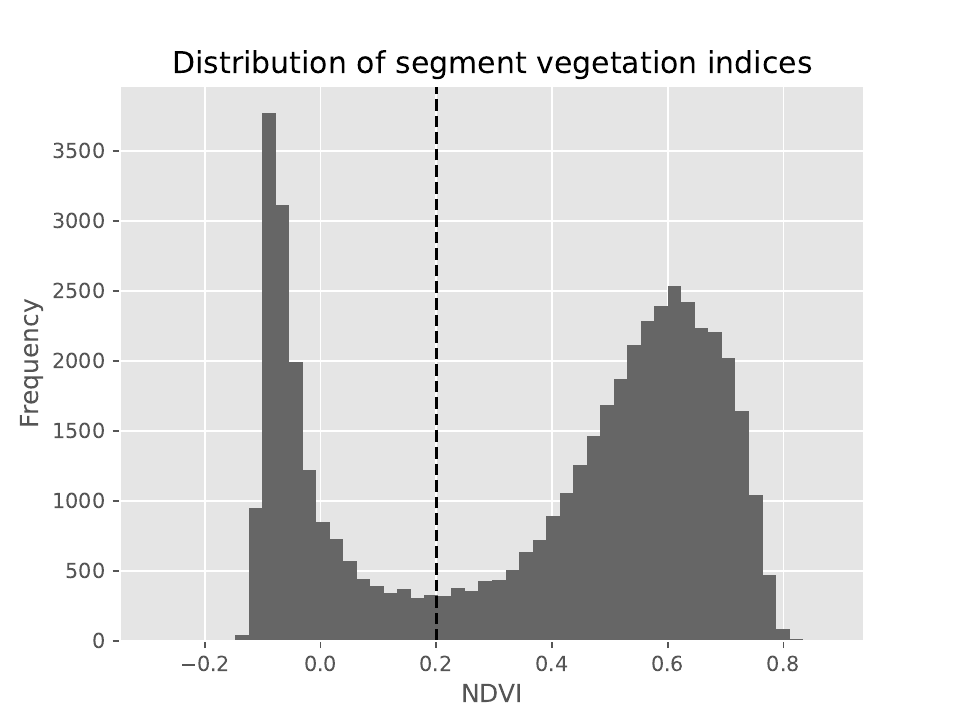}
    \caption{The RREI distribution of initial tree crown segments with the cut-off threshold indicated with the black dashed vertical line.}
    \label{fig:ndvi_hist}
\end{figure}

\clearpage

\begin{figure}[!ht]
    \includegraphics[width=\linewidth]{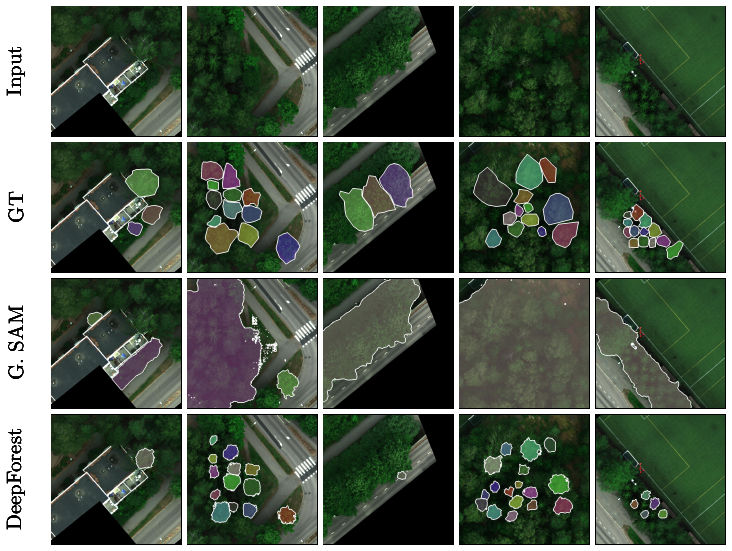}
    \caption{Visualisation of the outputs from the Grounded SAM (denoted G. SAM) and DeepForest models. The pseudo-randomly selected colours are used only to distinguish separate instance segments from each other. Each column represents a single test image with outputs from different models, with rows representing different models as indicated on the left-hand side.}
    \label{fig:main_visuals_app}
\end{figure}
\clearpage

\thispagestyle{empty}

\begin{figure}[!ht]
    \includegraphics[width=0.9\linewidth]{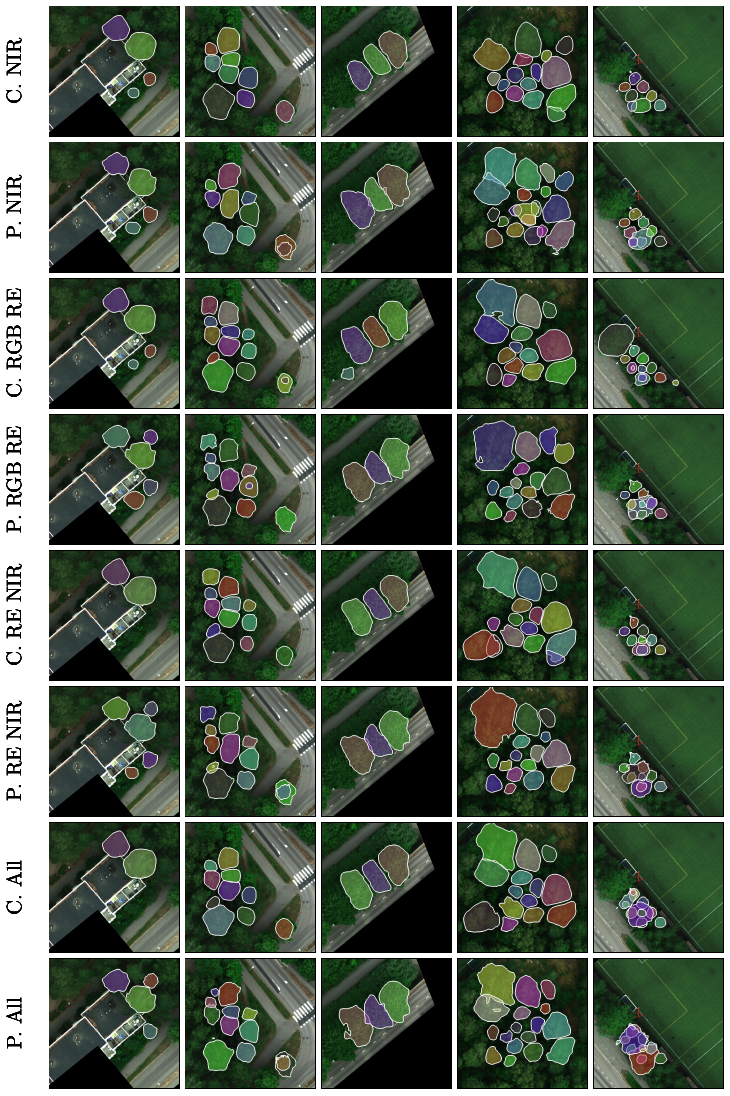}
    \caption{Visualisation of the outputs from models using different input modalities and training labels. C. is short for coarse, and P. for pseudo-labels, respectively. \textbf{All} refers to models using all three input modalities (RGB, RE, and NIR). The pseudo-randomly selected colours are used only to distinguish separate instance segments from each other. Each column represents a single test image with outputs from different models, with rows representing different models as indicated on the left-hand side.}
    \label{fig:abl_rest}
\end{figure}

\begin{figure}[!ht]
    \centering
    \includegraphics[width=\linewidth]{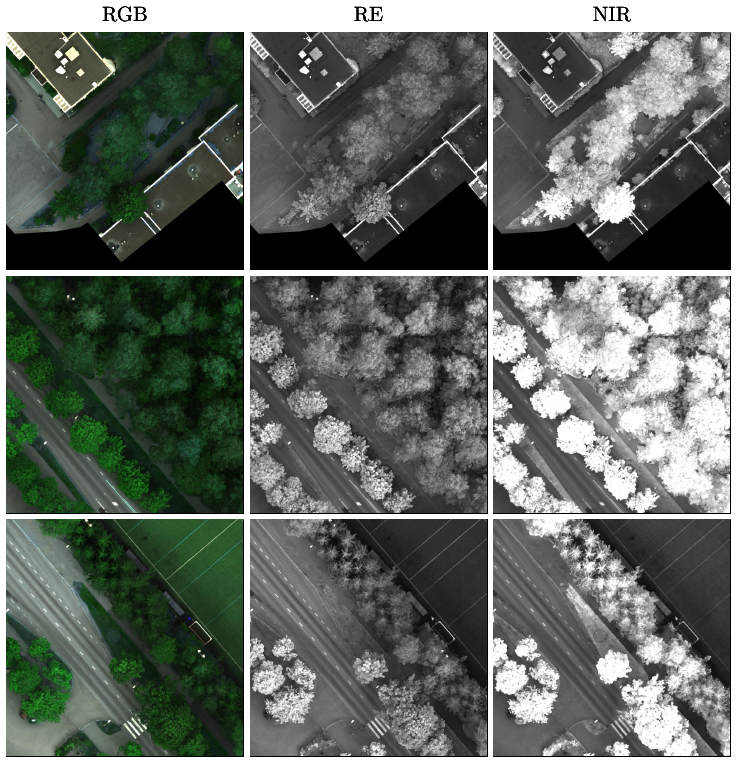}
    \caption{Visual comparison of unnormalised input channels as used by the proposed models. The samples were chosen to highlight how the trees are more difficult to distinguish from the background in certain RE images compared to the two other input modalities.}
    \label{fig:input_channel_comp}
\end{figure}
\clearpage

\section{Full Detectree2 Results}
\label{app:detectree2}
\setcounter{table}{0}

\begin{table}[!ht]
    \centering
    \caption{Test set metrics from all different Detectree2 model checkpoints.}
    \begin{tabular}{lrrrrr}
        \toprule
        \textbf{Detectree2 checkpoint} & \textbf{F1} & \textbf{Precision} & \textbf{Recall} & \textbf{mIoU} & \textbf{mIoU 95\% CI} \\
        \midrule
        \textit{base} & 0.298 & 0.540 & 0.206 & 0.182 & [0.150, 0.215] \\
        \textit{05dates} & 0.423 & 0.606 & 0.325 & 0.261 & [0.227, 0.298] \\
        \textit{flexi} & \textbf{0.556} & \textbf{0.911} & \textbf{0.400} & \textbf{0.324} & [0.281, 0.364] \\
        \textit{randresize\_full} & \underline{0.447} & \underline{0.718} & 0.325 & 0.269 & [0.228, 0.308] \\
        \textit{tropical\_closed\_canopy} & 0.064 & 0.177 & 0.039 & 0.039 & [0.025, 0.056] \\
        \textit{urban\_trees\_Cambridge} & 0.393 & 0.600 & 0.292 & 0.243 & [0.211, 0.276] \\
        \textit{withParacouUAV} & 0.415 & 0.543 & \underline{0.336} & \underline{0.280} & [0.242, 0.319] \\
        \bottomrule
    \end{tabular}
    \label{tab:detectree2}
\end{table}

\section{Evaluation near image edges}
\label{app:edge_eval}

\begin{table}[!ht]
    \centering
    \footnotesize
    \caption{Results when evaluating models based on full image predictions (including partial crowns closer to the image edges). All results are from the pseudo-supervised models.}
    \begin{tabular}{lrrr}
    \toprule
       \textbf{Input} & \textbf{mIoU} & \textbf{Change} & \textbf{Relative}(\%) \\
       \midrule
       RGB & 0.580 & -0.020 & -3.33 \\
       \textbf{RE} & \textbf{0.598} & +0.001 & 1.70 \\
       NIR & 0.553 & -0.035 & -5.97 \\
       RGB + RE & 0.587 & -0.024 & -3.93 \\
       \underline{RGB + NIR} & \underline{0.594} & -0.027 & -4.35 \\
       RE + NIR & 0.584 & -0.016 & -2.62 \\
       RGB + RE + NIR & 0.579 & -0.023 & -3.82 \\
       \bottomrule
    \end{tabular}
    \label{tab:results:full_stride}
\end{table}

\section{Näslund Funcion Fitting}
\label{app:naslund}

\begin{figure}[!ht]
    \centering
    \includegraphics[width=0.9\linewidth]{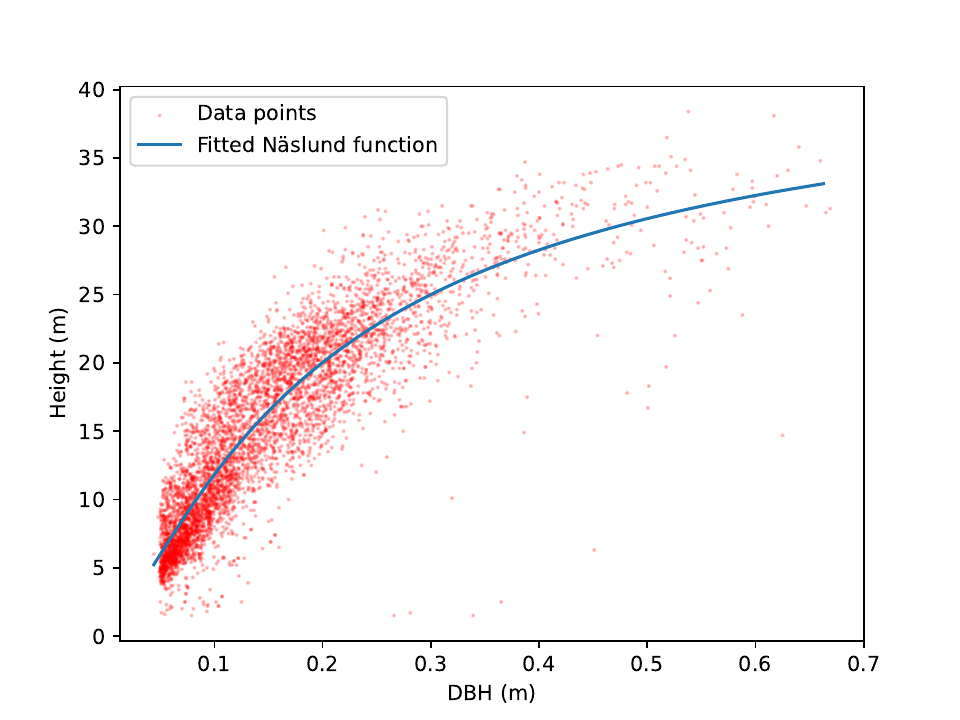}
    \caption{The fitted Näslund function in blue with the Evo DBH and tree height data points in red.}
    \label{fig:naslund}
\end{figure}

We fit the Näslund function:
\begin{equation}
    h = 1.3 \text{m} + \dfrac{DBH^2}{(aDBH + b)^2},
\end{equation}
where $h$ is the estimated tree height, DBH is the tree diameter at breast height, and $a$ and $b$ are the estimated coefficients, using data from Evo, Finland.
The data contained manual DBH and tree height from the years 2014, 2019, and 2021. 
There were a total of 2933 measured trees, of which 608 were Pinus sylvestris, 1283 were Picea abies, 734 were Betula sp., and the rest were of various deciduous species.
Between approximately 92 and 95\% of the trees were measured each year.
The tree diameter and height distribution is visualised as a scatter plot in Figure~\ref{fig:naslund} alongside the fitted Näslund function. 

The tree height estimates were not presumed to be too accurate, as they were simply used to aid in the categorisation of trees into different dominance classes based on their presumed sizes. 
Major error sources include the large geographical shift from Evo, Finland, to Sundsvall, Sweden and the fact that the model was simply fit to all measures instead of taking into account species-wise differences.

\end{document}